\documentclass[journal]{IEEEtran}
\usepackage{amsmath,amsfonts}
\usepackage[ruled,linesnumbered,vlined]{algorithm2e}
\SetKwInOut{Input}{Input}
\SetKwInOut{Output}{Output}
\SetKw{Continue}{continue}%
\SetKw{Break}{break}%
\SetKw{KwAnd}{and}%
\SetKw{KwOr}{or}%
\SetKw{KwNot}{not}%
\usepackage{array}
\usepackage[caption=false,font=footnotesize]{subfig}
\usepackage{textcomp}
\usepackage{stfloats}
\usepackage{url}
\usepackage{verbatim}
\usepackage{graphicx}
\usepackage{cite}
\usepackage{multirow}
\usepackage{makecell}
\usepackage{multicol}
\usepackage{booktabs}
\usepackage{soul}
\usepackage{amsthm,amssymb}

\usepackage{xcolor}
\usepackage{hyperref}
\usepackage[capitalise,nameinlink]{cleveref}

\definecolor{mLinkBlue}{HTML}{002d7d}
\hypersetup{
  colorlinks,
  linkcolor={mLinkBlue},
  citecolor={mLinkBlue},
  urlcolor={mLinkBlue}
}
\newcommand{\mc}{\mathcal}
\newcommand{\mb}[1]{{\mathbf{#1}}}

\definecolor{mBlue}{HTML}{1F77B4}
\definecolor{mDarkRed}{HTML}{a2282f}

\newcommand*{\Mbar}{\overline{\mc{M}}}
\newcommand*{\Nin}[1][i]{\mc{N}_{#1}^{\mathrm{in}}}
\newcommand*{\Nout}[1][i]{\mc{N}_{#1}^{\mathrm{out}}}
\newcommand*{\fin}[1][i]{{f}_{#1}^{\mathrm{in}}}

\newcommand*{\bfout}[1][i]{\mb{f}_{#1}^{\mathrm{out}}}

\newtheorem{definition}{Definition}
\newtheorem*{remark}{Remark}
\newtheorem{proposition}{Proposition}

\newcommand{\fgref}[1]{\cref{#1}}
\newcommand{\scref}[1]{\Cref{#1}}
\renewcommand{\eqref}[1]{\labelcref{#1}}
\graphicspath{{./files/figures/}}

\usepackage{pgfplots}
\pgfplotsset{compat=newest}
\definecolor{mLightGreen}{HTML}{14B03D}
\definecolor{mDarkRed}{HTML}{a2282f}
\tikzstyle{mydashdot}=[dash pattern=on 6pt off 2pt on \the\pgflinewidth off 2pt]
\usetikzlibrary{matrix}
\usepgfplotslibrary{groupplots}
\pgfplotsset{every tick label/.append style={font=\tiny}}
\pgfplotsset{ylabsh/.style={every axis y label/.style={at={(0,0.5)}, xshift=#1, rotate=90}}}

\begin{document}
\bstctlcite{IEEEexample:BSTcontrol}
\title{Online Multi-Robot Coordination and Cooperation with Task Precedence Relationships}
\author{Walker Gosrich, Saurav Agarwal, Kashish Garg, Siddharth Mayya,\\
  Matthew Malencia, Mark Yim, and Vijay Kumar
  \thanks{The work was supported by ARL DCIST CRA W911NF-17-2-0181 and NSF Award 2415249. This material is based upon work supported by the National Science Foundation Graduate Research Fellowship.}%
  \thanks{W.\@ Gosrich, S.\@ Agarwal, K.\@ Garg, M.\@ Yim, and V.\@ Kumar are with the GRASP Laboratory, University of Pennsylvania, Philadelphia, PA, USA. {\{\href{mailto:gosrich@seas.upenn.edu}{gosrich}, \href{mailto:sauravag@seas.upenn.edu}{sauravag}, \href{mailto:kashg@seas.upenn.edu}{kashg}, \href{mailto:yim@seas.upenn.edu}{yim}, \href{mailto:kumar@seas.upenn.edu}{kumar}\}@seas.upenn.edu }}
  \thanks{S.\@ Mayya is with Amazon Robotics, North Reading, MA, USA.}%
  \thanks{M.\@ Malencia is with Zipline, San Francisco, CA, USA.}%
  \thanks{This work is not related to Amazon or Zipline.}%
}

\maketitle

\begin{abstract}
  We propose a new formulation for the multi-robot task allocation problem that incorporates (a)~complex precedence relationships between tasks, (b)~efficient intra-task coordination, and (c)~cooperation through the formation of robot coalitions.
  A task graph specifies the tasks and their relationships, and a set of reward functions models the effects of coalition size and preceding task performance.
  Maximizing task rewards is NP-hard; hence, we propose network flow-based algorithms to approximate solutions efficiently.
  A novel online algorithm performs iterative re-allocation, providing robustness to task failures and model inaccuracies to achieve higher performance than offline approaches.
  We comprehensively evaluate the algorithms in a testbed with random missions and reward functions and compare them to a mixed-integer solver and a greedy heuristic.
  Additionally, we validate the overall approach in an advanced simulator, modeling reward functions based on realistic physical phenomena and executing the tasks with realistic robot dynamics.
  Results establish efficacy in modeling complex missions and efficiency in generating high-fidelity task plans while leveraging task relationships.
\end{abstract}


\section{Introduction}\label{sec:intro}
\begin{figure}[t]
	\centering
	\includegraphics[width=\columnwidth]{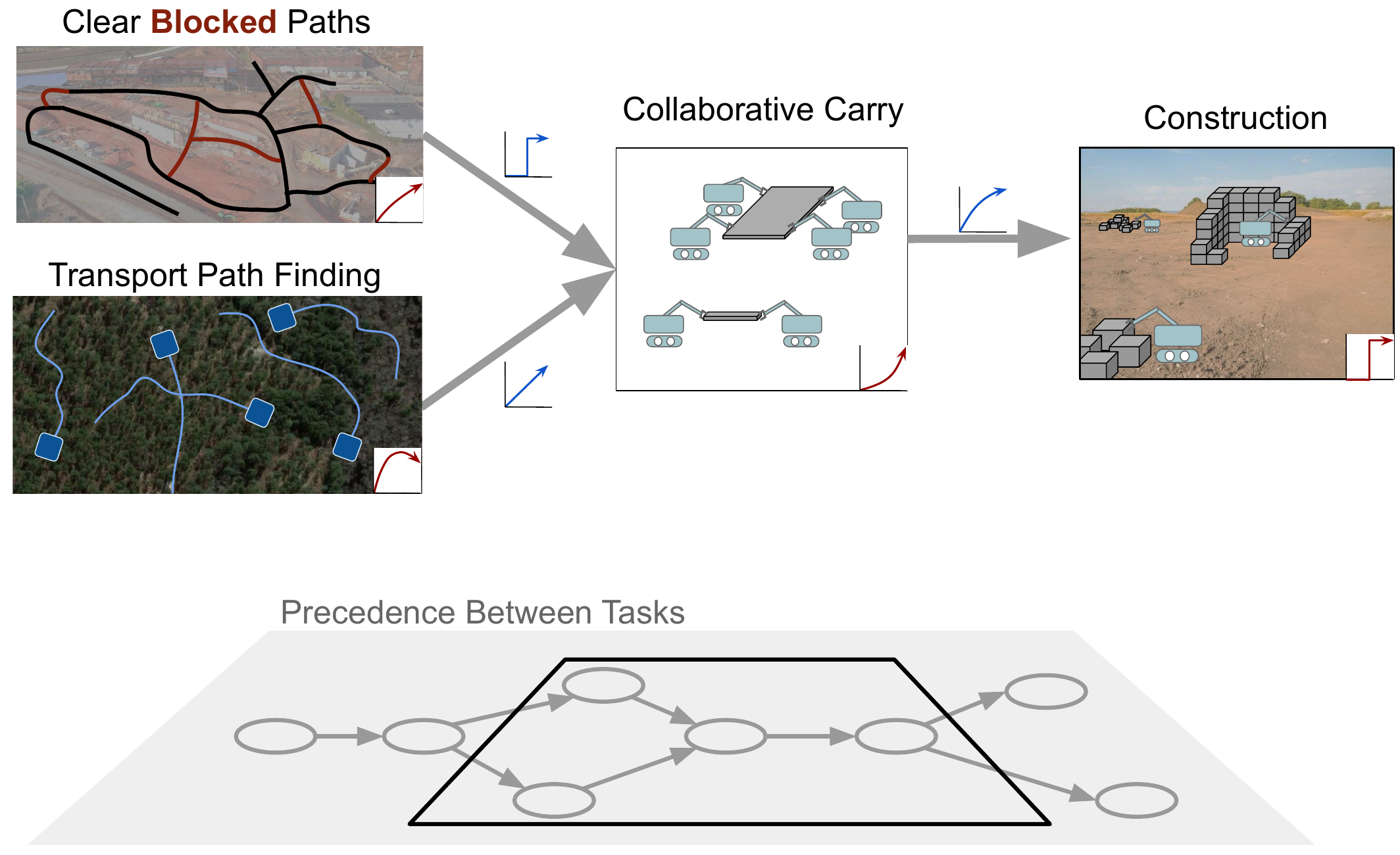}
  \caption{\label{fig:intro_illustration}%
  The task graph (\textit{bottom}) shows precedence relationships between tasks in a mission.
  Highlighted is a subset of tasks, showing coalition functions (\textcolor{mDarkRed}{\textit{red}}) on task nodes and precedence relationships (\textcolor{mBlue}{\textit{blue}}) on edges.
For example, before transporting construction materials, robots may need to find or clear paths through a debris-filled construction site.
The quality of these paths directly impacts the team's ability to transport material.
Furthermore, the performance of transporting materials, e.g., whether materials get damaged, impacts the quality and speed of a following construction task.}
\end{figure}

\IEEEPARstart{L}{arge}-scale missions often comprise several tasks that have \textit{precedence relationships} among them, i.e., the quality of a task's execution may depend on how well a related task is executed, which need not be a simple binary relationship.
	Additionally, task outcomes are often uncertain products of complex interactions between robots and the environment.
As an example, the illustration in \fgref{fig:intro_illustration} shows an autonomous construction mission where transport paths must be identified and cleared, and materials transported before construction takes place.
In such a scenario, the quality with which prior tasks are completed impacts the ability of a robot coalition to complete future tasks, e.g., well-cleared paths will permit more agents to transport materials through.
Furthermore, tasks such as collaborative carry require multiple robots to work together to complete the task when a single robot cannot, whereas other tasks, such as exploration for transport pathfinding, may see improved results from multiple robots working together.

A fundamental challenge in such missions is to efficiently execute the tasks to maximize the cumulative reward, while respecting additional resource constraints such as time and robot availability.
This optimization problem is NP-hard and belongs to a broader class of multi-robot task allocation (MRTA) problems.
Approaches for such problems must compute robot \textit{coalitions} to collaboratively execute individual tasks and an allocation of these coalitions to various tasks.
Accomplishing this requires a model of the expected performance of a given coalition assigned to a task, and a model of the relationships among tasks, especially the \textit{influence} of a preceding task on subsequent tasks.
The complex relationships among tasks and robots are critical in real-world applications, but have not been addressed in the literature in a unified manner, owing to the modeling complexity and computational challenges.

The second fundamental challenge in executing a mission is uncertainty due to the dynamic nature of the environment, inaccuracies in the mission model due to sim-to-real gaps, and robot failures.
While an offline solution can be computed before the mission execution, it may not be robust to these uncertainties.
On the other hand, an online solver that can re-plan in response to actual task outcomes is more robust to uncertainties and failures.
For example, a task~B may be highly dependent on the quality of the execution of a preceding task~A. If task~A fails or performs poorly, it may be advantageous to reallocate robot resources away from task~B.
However, as the problem is NP-hard, an optimal approach such as a mixed-integer non-linear program (MINLP) can be computationally expensive to execute in real time after every individual task completion.
Hence, an online algorithm that effectively trades off solution quality and computational complexity for real-time execution while being adaptive to uncertainties and failures is desirable.

This article presents a novel formulation for modeling large-scale multi-robot missions via precedence relationships among tasks, as well as coordination and cooperation among robots through coalitions.
We present an online iterative algorithmic framework that leverages this model for allocating coalitions to tasks and adapting to task failures and uncertainty.
The formulation is applicable to complex scenarios relevant to autonomous construction, precision agriculture, and industrial robotic applications.
The \textit{contributions} are:
\begin{enumerate}
	\item A modular and expressive mission model defined over a task graph and a reward function (\scref{sec:mission-model}).
	      The task graph captures the precedence relationships among tasks.
	      The reward function quantifies task utilities and is characterized by \textit{influence functions}---modeling of relationships among tasks, \textit{aggregation functions}---effect of all preceding tasks on a dependent task, and \textit{coalition functions}---modeling of task execution efficacies with robot coalition size.
	\item A non-linear network flow formulation along with offline and online iterative algorithms that efficiently form coalitions and allocate them to tasks while leveraging the task precedence relationships (\scref{sec:flow-solution}).
	      The online algorithm enables replanning in response to actual task outcomes, making it adaptive to task failures and inaccuracies in the mission model.
	\item A new testbed for mission models to randomly generate task graphs with complex rewards and precedence relationships, which facilitates extensive evaluation of the proposed algorithms.
	      The results show that the online algorithm computes solutions close to optimal (determined via the MINLP solver run for 12 hours) for small problem sizes and significantly outperforms baselines for large problem sizes with and without stochastic task failures and model inaccuracies.
	\item A Unity-based testbed with four canonical tasks---coverage control, exploration, resource transportation, and collaborative carry---to form large-scale missions in realistic urban environments with large robot teams (\scref{sec:advanced-experiments}).
		      Test missions of varying sizes and numbers of agents show that the proposed formulation is able to accurately ($<$20\% error) model task rewards in complex scenarios, and the online solver is able to leverage task reward feedback to provide robust mission performance.
\end{enumerate}

This article extends our work in~\cite{gosrich2023multi}, which introduced the task graph mission model, greedy and non-linear programming (NLP) flow-based solution methods, and a mixed integer non-linear programming (MINLP) solution method.
In particular, this article introduces an online iterative flow-based algorithm that performs on-the-fly replanning to account for updated observations from the environment.
We perform extensive simulation studies to establish the effectiveness of the proposed online algorithm in comparison to the offline approach proposed in~\cite{gosrich2023multi}.
Furthermore, we experimentally validate the modeling approach on a high-fidelity simulator of an urban environment with teams of up to 100 agents and mission sizes of up to 25 tasks.
Experimental validation includes the implementation of four tasks---coverage control, exploration, resource transportation, and collaborative carry---and the simulation of aerial robots performing these tasks.

\section{Background}
This article focuses on multi-task missions composed of multi-robot tasks with inter-task precedence relationships and uncertainty in task performance.
In applications where these mission characteristics are common, such as autonomous construction and assembly~\cite{knepper2013ikeabot}, agriculture~\cite{mao2021research}, and complex multi-robot missions, efficient allocation of robot coalitions that enables \textit{coordination}, \textit{cooperation}, and uncertainty tolerance is essential.
In prior work, these characteristics have been considered extensively, but rarely together in a cohesive model.

\subsection{Task Planning with Single-Robot Interrelated Tasks}
Many approaches allocate robots among interrelated tasks without considering multi-robot coalitions.
Some formulate the problem purely as a constraint satisfaction problem (CSP)~\cite{deng2019compiler, dogar2019multi}, others hybridize mixed integer linear programming (MILP) with CSP in order to reason about task utility~\cite{gombolay2013fast}.
In operations research~\cite{NUNES2017}, the task allocation problem with inter-task relationships is similar to the Vehicle Routing Problem with Time Windows~\cite{kolen1987vehicle}, with additional ordering constraints~\cite{bredstrom2008combined}.
However, these do not consider multi-agent tasks or expressive inter-task relationships.

In the AI planning literature, multi-agent planning (MAP) with \textit{joint actions} expresses the tight coupling between tasks that we represent with precedence relationships.
Several approaches~\cite{brafman2014distributed, shekhar2020signaling} consider multi-agent tasks and tight task coupling, but only consider coalitions in a limited fashion.
Some approaches~\cite{TERESHCHUK2021102154, smith2019real, ponda2010decentralized} consider soft precedence constraints---precedence constraints that degrade reward when violated.
Other methods~\cite{smith2019real, wang2022heterogeneous} model inter-task relationships among single-agent tasks in detail on a heterogeneous graph, with nodes representing tasks, agents, and locations.

\subsection{Task Planning with Cooperation}
Some approaches focus on \textit{cooperation}, modeling the relationship between the coalition assigned to a task and task performance.
These approaches model reward as a function of the number of homogeneous robots assigned to complete the task~\cite{Korsah2013, seenu2020review}, as a function of heterogeneous robot traits assigned to the task~\cite{prorok2017impact}, or do not consider task reward explicitly~\cite{messing2022grstaps}.
 
Other approaches model agents by assigning a population fraction---a proportion of the total robot team---to \textit{flow} between tasks, scaling well to large teams.
Some of these approaches require a task allocation distribution known a priori~\cite{hsieh2008biologically, berman2009opti}, while others consider the heterogeneous case with simplified reward models and no task relationships~\cite{solovey2021fast}.

\subsection{Task Planning in Uncertain Environments}
The two primary approaches for uncertain environments are explicitly modeling uncertainty and online reactive reallocation to handle unexpected events and observations.

Explicit modeling of uncertainty is powerful when enough data is present.
In homogeneous agent problems, some approaches address multi-agent tasks using redundant assignment via optimization to reduce risk~\cite{prorok2020robust, malencia2021fair}.
Others address assignment to single-agent tasks using the conditional value at risk (CVaR) metric~\cite{nam2016analyzing} and chance-constrained optimization~\cite{ponda2012distributed}.
Allocation approaches with heterogeneous agents model uncertainty in task requirements, robot capabilities, task duration, or inter-task travel times.
Approaches leverage the CVaR cost~\cite{fu2022robust}, min-max optimization of success probability~\cite{rudolph2021desperate}, or the sequential probability ratio test~\cite{messing2023sampling, park2023risk-tolerant}.
These methods are promising for reducing risk, but none consider complex inter-task relationships, and many consider only simple linear reward models.

Another approach is online reallocation to handle unexpected events. 
This approach is rooted in spatial task allocation and formation control problems, which frequently employ gradient-based~\cite{schwager2011unifying} or market-based assignment and control~\cite{michael2008distributed}. Spatial task allocation is flexible and reactive, able to trade off dynamic and heterogeneous tasks and handle noise~\cite{amir2022multiagentdistributeddecentralizedgeometric}.
In the homogeneous agent setting with single-agent tasks and temporal constraints, online approaches typically address emerging dynamic tasks by instantaneous assignment~\cite{nunes2015multi, GHASSEMI2022103905, johnson2016decentralized}.
In heterogeneous agent settings with multi-robot tasks, some methods perform short horizon task assignment~\cite{mayya2021resilient} without considering complex inter-task relationships, or perform dynamic repair of time-extended assignments~\cite{neville2023d} without considering task reward.

\subsection{Task Specification}
The field of task specification addresses the translation of abstract mission representations into precise mathematical encodings, often using linear temporal logic (LTL)~\cite{bai2021multi, SCHILLINGER2021103866} or natural language.
The task specification problem often requires domain-specific approaches~\cite{nordmann2014survey}.
In this work, we assume access to a well-defined task specification; this effectively decouples our task allocation method from the task specification mechanism.
We describe in~\scref{sec:mission-modeling-process} and demonstrate in~\scref{sec:advanced-experiments} the translation of an informal mission specification into one that is useful for this solver. 

Recently, the use of large language models (LLMs) for task specification, allocation, and planning has gained popularity, due to the ease of specifying robot capabilities and task contexts~\cite{pmlr-v205-ichter23a}.
These approaches typically handle task planning with single robots~\cite{singh2023} or address multiple agents on a single complex task~\cite{hong2023metagpt}.
Notably,~\cite{kannan2024smartllmsmartmultiagentrobot} addresses multi-agent tasks with task relationships, though only on small problems (a few tasks and robots), and uses task decomposition paired with a binary reward model.
LLMs are promising for task \textit{specification} and translation into a problem definition, generating code to execute tasks, and in-the-loop replanning based on natural language feedback.
However, LLMs presently lack the ability to generate high-quality solutions for complex problems.

\subsection{Unifying Coordination, Cooperation, and Uncertainty}
Some approaches, like ours, consider the challenging intersection of these problem characteristics, requiring a system to reason about intersecting robot schedules and fluidly form coalitions among different tasks over an extended time horizon.
The approaches that consider both task order and coalition typically use binary task rewards~\cite{capezzuto2020anytime, suslova2020multi, ramchurn2010coalition, neville2024qitagsqualityoptimizedspatiotemporalheterogeneous}, and some cannot scale to problems larger than a few robots and tasks~\cite{korsah2012xbots}.
None of these approaches consider uncertainty.
In this work, we present an expressive model for task reward that expands these models, solution methods that leverage the model to scale to large problem sizes (50+ tasks, 100 agents), and an online reactive algorithm capable of re-planning in uncertain environments.

\section{Mission Model}\label{sec:mission-model}
In this section, we introduce \textit{task graphs}, which encode the structure of task relationships, and we define a \textit{task reward model} over the task graph.
Taken together, these models can represent complex missions for homogeneous robot teams.

Let $\mc{T} :=\{T_1,\ldots,T_M\}$ represent the set of $M$ available tasks in the environment, and $\mc{M} :=\{1,\ldots,M\}$ represent the index set of tasks.
Each task~$T_j$ has a constant duration~$d_j$ and yields a reward~$r_j$ upon completion.
The reward is a function of the assigned robot coalition and the rewards of related preceding tasks, as defined later in the section.

\begin{definition}[Task Graph]%
	Given a set of tasks $\mc{T}$, the task graph is a directed acyclic graph $\mc{G}_T = (\Mbar, \mc{E})$, where $\Mbar = \{0,1,\ldots,M\}$ is the vertex set corresponding to tasks and $\mc{E}$ is the edge set corresponding to task relationships.
	The vertex~$0$ represents a virtual source node, and all other vertices correspond to actual tasks.
	We add a directed edge $(i, j)$ to the set $\mc E$ if there exists a precedence relationship between tasks $T_i$ and $T_j$ such that the reward $r_i$ associated with task $T_i$ impacts the reward $r_j$ from task $T_j$.
	We add an edge from the source node $T_0$ to all tasks without preceding neighbors, resulting in a directed acyclic graph.
	We associate an inter-task travel time $\Delta_{ij} \in \mathbb{R}^+ \quad \forall i,j \in \mc{M}$ between tasks $T_i$ and $T_j$ with the assumption that $\Delta_{ii}=0$.
\end{definition}%

Let $\mc R$ represent the index set of $N$ robots available to execute the tasks, and let $x_k^r \in \{0, 1\}$ indicate whether a robot $r\in \mc R$ executes task~$k\in \overline{\mathcal M}$ during the mission.
We define a robot coalition $\mc{C}_j$ as the set of robots allocated to task $T_j$.
Finally, let $C_j = |\mc{C}_j|$ represent the size of the coalition, where $|\cdot|$ is the set cardinality operator.

\subsection{Task Reward Model}\label{sec:task_model}
We present a model for the reward $r_j$, associated with the task $T_j\in \mc T$.
The task reward model comprises three functions:
1)~a \textit{coalition function} $\rho_j$;
2)~\textit{influence functions} $\delta_{ij}$; and
3)~an \textit{aggregation function} $\alpha_j$.

\begin{definition}[Task Coalition Function]%
	Given a robot coalition $\mc{C}_j$ assigned to task $T_j$, the task coalition function $\rho_j(C_j): \mathbb{R} \mapsto \mathbb{R}$ returns a scalar that represents the effectiveness of the robot coalition at accomplishing the given task.
\end{definition}

\begin{figure}[tbp]%
	\centering
	\includegraphics[width=\columnwidth]{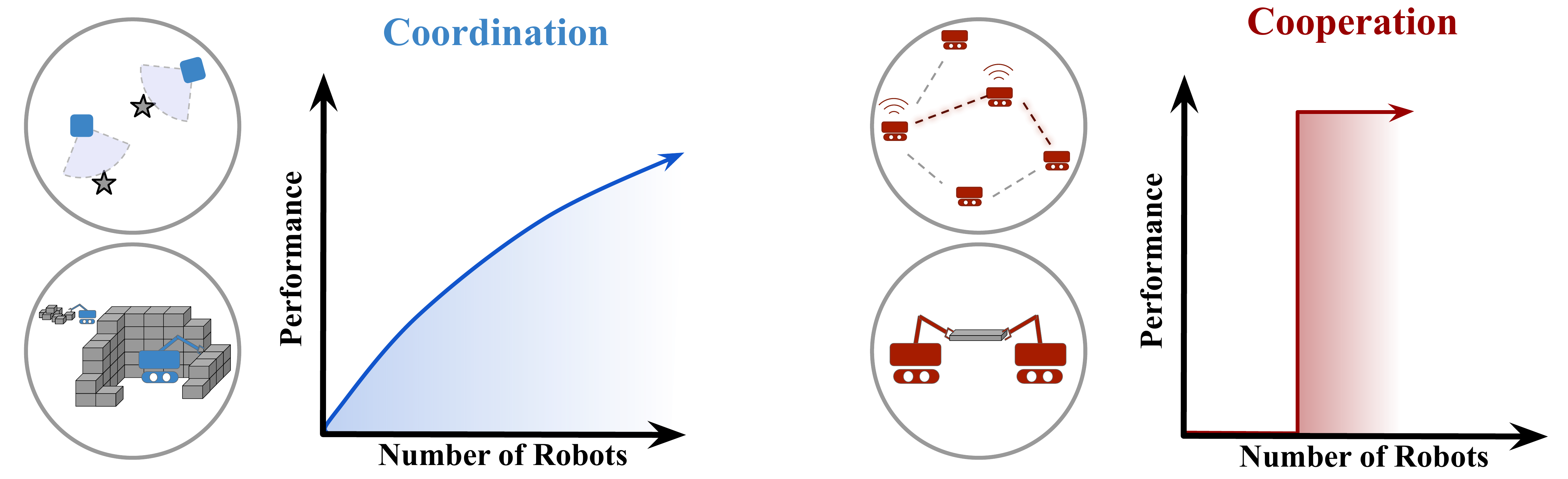}
	\caption{\label{fig:coalition}%
		Figure adapted from~\cite{prorok2021beyond}.
		Two types of coalition function $\rho$.
		\textit{Left}: Tasks such as coverage control and construction have a subadditive coalition function, as larger coalitions improve performance.
		\textit{Right}: Network connectivity tasks and cooperative transport are represented by a step coalition function, where a critical number of robots is required to perform the task.%
	}
\end{figure}

\begin{figure}[tbp]%
	\centering
	\includegraphics[trim={0.0cm 1.7cm 0.0cm 0.40cm},clip,width=\columnwidth]{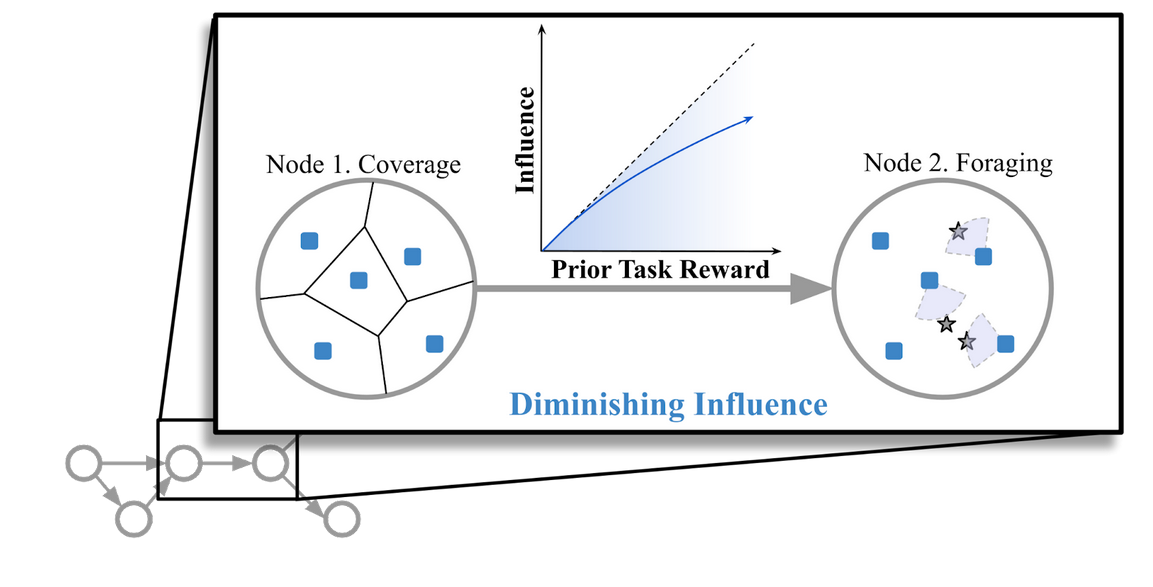}
	\caption{\label{fig:influence}%
		The influence function, illustrated here, models the relationship between the reward at the preceding (or \textit{influencing}) task and the reward at the current task.
		A better result on a coverage task would result in higher performance in the foraging task (exploiting the information accrued during coverage).
		This is modeled by a subadditive influence function.}%
\end{figure}

The coalition function provides an expressive model for the relationship between the robot coalitions assigned to a task and the resulting task reward.
While prior works~\cite{dutta2019one,zitouni2019fa} implicitly assume that the coalition functions are linear, this article uses task coalition functions to model both linear and non-linear performance characteristics.
Examples of these more complex coalition-reward relationships are seen in~\fgref{fig:coalition} with common robotic tasks such as collaborative tasks with subadditive reward (left) and cooperative tasks that can only be achieved by a threshold number of robots (right).

We now model the impact of the precedence relationships on the reward $r_j$ for each task $T_j$.
\begin{definition}[Task Influence Function]%
	The influence function $\delta_{ij} (r_i): \mathbb{R} \mapsto \mathbb{R},\; (i,j)\in\mathcal{E}$ returns a scalar that quantifies the extent to which the performance of the robots on task $T_i$ hinders or facilitates the execution of task $T_j$.
	In other words, it models the influence of task reward $r_i$ on $r_j$.
\end{definition}%

The task influence function provides a generalization of precedence constraints commonly studied in task allocation models.
Typical approaches model task precedence as a binary construct: a dependent task may be executed only if a preceding task is completed with performance above a certain threshold.
In contrast, representing $\delta_{ij}$ as a function enables the modeling of complex relationships between tasks often found in real-world scenarios.
For instance, as shown in~\fgref{fig:influence}, the performance of an exploration or a coverage task could be sub-linearly related to the performance of a subsequent transportation or foraging task.
Similarly, the performance of a foundation leveling task could influence future construction tasks on top of that foundation.
Furthermore, our formulation can also represent a classical binary precedence constraint by choosing $\delta_{ij}$ as a step function.

When a task $T_j$ has multiple incident edges, the outputs of these multiple influence functions must be aggregated over the set of incoming precedence tasks $\Nin[j]= \{T_i\mid (i,j) \in \mc{E}\}$.
\begin{definition}[Task Influence Aggregation Function]%
	The task influence aggregation function $\alpha_j: \mathbb{R}^{\lvert\Nin[j]\rvert} \mapsto \mathbb{R}$ takes as an input the set of influencing rewards $ \{r_i \mid i \in \Nin[j]\}$ and outputs a value representing the total influence of preceding tasks on the reward $r_j$ associated with task $T_j$.
\end{definition}

The task influence aggregation function can be set to any arbitrary function in $\mathbb{R}^{\lvert\Nin[j]\rvert} \mapsto \mathbb{R}$, such as $\sum_{i \in \Nin[j]}\delta_{ij}(r_i)$ to represent the case where completing any preceding task can result in a reward on task $T_j$ (logical ``OR''), or $\prod_{i \in \Nin[j]}\delta_{ij}(r_i)$ to represent the case when all preceding tasks must be sufficiently completed to achieve favorable performance on task $T_j$ (logical ``AND'').

The reward $r_j$ accrued by each task $T_j$ represents the quality of the outcome of the task.
\begin{equation}%
	\label{eq:reward_dynamics}
	r_j = \rho_j(C_j) \; \ddag_j \; \alpha_j\left(\{\delta_{ij}(r_i)\mid i \in \Nin[j]\}\right)
\end{equation}
In the context of task $T_j$, the symbol $\ddag_j$ represents a function that combines outputs of the coalition function $\rho_j$ and aggregated influence function $\alpha_j$.
For example, summing $\rho_j$ and $\alpha_j$ can represent a ``soft'' constraint, where the reward is non-zero even if precedence-related tasks are incomplete.
Alternatively, the product of $\rho_j$ and $\alpha_j$ could be interpreted as a ``hard'' constraint, where a zero aggregated influence value results in a task reward of zero.

Taken together, the task graph $\mc{G}_T$, the set of coalition, aggregation, and $\ddag$ functions on its vertices, $\{\rho_j, \alpha_j, \ddag_j \}, \forall j\in\mc{M}$, and the set of influence functions defined over its edges $\{\delta_{ij}\}, \forall (i, j)\in\mc{E}$ completely specify the mission.

\subsection{Problem Statement, Characteristics, and Assumptions}\label{subsec:ps}
The objective of the MRTA problem is to generate a set of feasible robot-task assignments $x^r_k, \forall r\in\mc{R}, k\in\mc{M}$, forming a set of coalitions $\mc{C}$ that maximizes the rewards obtained over all tasks:
\begin{equation}%
	\label{eq:cost}
	\max_{\mc{C}_j, \forall j\in \mc{M}} \sum_{j=1}^{M} r_j
\end{equation}
while satisfying the following constraints:
\textit{(i)}~the last task is completed by the robots before the makespan time~$\tau$,
\textit{(ii)}~robots in a coalition work on only one task at a time,
\textit{(iii)}~the robot-task assignments respect the precedence relationships defined over the graph edges $(i,j) \in \mc{E}$,
\textit{(iv)}~the task coalitions do not exceed the total number of robots~$N$,
\textit{(v)}~tasks commence only when all assigned robots have arrived at the task location, and
\textit{(vi)}~once a coalition of robots starts a task~$i$, it continues working on the task for its entire duration~$d_i$.
We assume non-negative reward functions.
See \scref{sec:minlp,sec:flow-solution} for specific instantiations of these constraints.

\begin{proposition}
  There is no $\alpha$-approximation algorithm for any constant $\alpha\leq 1$ for the MRTA problem unless \emph{P$=$NP}.
\end{proposition}
\begin{proof}
We perform a \textit{gap-preserving reduction} from the maximization version of the $d$-regular label cover problem defined on a bipartite graph $G=(U, V, E)$ with label sets $L_u$ and $L_v$ for $U$ and $V$, respectively~\cite{WilliamsonApproximationBook}.
For each edge $(u, v) \in E$, $R_{(u,v)}\subseteq L_u \times L_v$ defines a non-empty relation of acceptable labels, i.e., the edge is satisfied if the vertices $u\in U$ and $v\in V$ are assigned labels $l_u$ and $l_v$ such that $(l_u, l_v)\in R_{(u,v)}$. In a $d$-regular instance, each vertex has a degree $d$, and $\vert U\vert = \vert V\vert = n$.
Let $m$ be the number of edges.

We now reduce an instance $I$ of the $d$-regular label cover problem to an instance of the MRTA problem.
For every vertex $u\in U$ and $v\in V$, create a \textit{source task}.
For every pair of vertex and label $(u, l)$, where $l\in L_u$, create a \textit{labeled task}.
Similarly, create a labeled task for each pair $(v, l)$, where $l\in L_v$.
For each source and labeled task, the coalition function (and reward function) returns $\epsilon >0$ if the task is executed by exactly one robot and 0 otherwise.
We add a precedence relationship for each labeled task to its corresponding source task $s$ with reward $r_s$.
The task influence function is defined as $\mathbb{I}[r_s > 0]$, where $\mathbb I$ is the Iverson bracket.
The combination function $\ddag$ is a product, so that a labeled task gets a reward of $\epsilon$ if and only if both the labeled task and its corresponding source task are executed by exactly one robot.

Now, add an \textit{edge task} for each relation $(u_l, v_l)\in R_{(u, v)}$ of each edge $(u, v)\in E$.
The coalition function returns 1 if the task is executed by 2 robots and 0 otherwise.
We add a preceding relationship edge to labeled tasks $u_l$ and $v_l$ with task influence functions $\mathbb{I}[r_{u_l} > 0]$ and $\mathbb{I}[r_{v_l} > 0]$.
The influence aggregation and the combination functions are products.
Thus, an edge task returns a reward of~1 if and only if it is executed by 2 robots, and both the corresponding labeled tasks must have been executed by~1 robot each.
This inherently means that the corresponding source tasks must also be executed by~1 robot each.

Finally, we assign a task duration of 1 to each task and a makespan time limit of 3. We supply $2n$ robots.
We set $\epsilon$ to a very small positive value such that $6\epsilon n \ll 1$.
The maximum reward that an optimal solver can get is when it executes all edge tasks.
In doing so, it will ensure the execution of all source tasks and one labeled task per source, resulting in a total reward of $4\epsilon n +m$.
The edge tasks executed by the solution correspond to the edges satisfied in the label cover problem.
For the label cover problem, it is NP-hard to distinguish between an instance where $m$ edges can be satisfied and instances where at most $\alpha m$ edges can be satisfied, for any $\alpha < 1$~\cite{WilliamsonApproximationBook}.
An $\alpha + \epsilon'$ approximation algorithm, with arbitrarily small positive $\epsilon'$, for MRTA would imply P$=$NP, as it will be able to distinguish between the two instances of the label cover problem.
Hence, it is NP-hard to approximate the MRTA problem within any constant factor.
\end{proof}

\subsection{Mixed Integer Non-Linear Program (MINLP)}
\label{sec:minlp}
We now describe an MINLP formulation that leverages the mission model to pose the task allocation problem as an optimization problem.
The objective is to maximize the total reward, given in~\eqref{eq:reward_dynamics} and~\eqref{eq:cost}, with the coalition size of robots assigned to a task $T_k,\ k\in \mc M$ given by $C_k = \sum_{r\in \mc R}x^r_k$.

We let non-negative continuous variables $S_k$ and $F_k$ denote the start and the finish times for the task~$T_k$.
Additionally, let the binary variable $x_k$ be non-zero when the task $T_k$ is executed by at least one robot, as ensured by constraints~\eqref{eqn:oneTask} below.
The precedence relationships are dictated by the task graph $\mc G_T=(\overline{\mc M}, \mc E)$.
The following constraints~\eqref{eqn:precedence} and~\eqref{eqn:qij}, with the help of auxiliary binary variables $w_{ij}$, ensure that for an edge $(i, j) \in \mc E$, the preceding task $T_i$ is executed before the dependent task $T_j$, if and only if both $T_i$ and $T_j$ are executed.
Furthermore, for each executed task, we need to ensure that the duration constraints~\eqref{eqn:duration} are enforced.
\begin{align}
	 & x_k \geq x_k^r, \, \forall r\in \mc R \text{ and } x_k \leq C_k, \quad \forall k \in \mc M\label{eqn:oneTask} \\
	 & w_{ij} (S_j - F_i) \geq 0,\; w_{ij}\geq x_i + x_j -1,\label{eqn:precedence}                                                      \\
	 & w_{ij}\leq x_i,\;w_{ij} \leq x_j,\text{ and } w_{ij}\in\{0,1\},\quad \forall (i, j) \in \mc E\label{eqn:qij}                     \\
	 & x_k \left(F_k - S_k\right) \geq x_k d_k, \quad \forall k \in \mc M\label{eqn:duration}
\end{align}

The rest of the constraints follow from the formulation by Nunes et al.~\cite{NUNES2017}.
Binary variables $z_k^r$ denote if $k$ is the last task executed by the robot~$r$, and binary variables $o_{kk'}^r$ denote if robot~$r$ executes a task $k'\in \mc M$ immediately after task $k$.
Inter-task travel time $\Delta_{ij}$ is incorporated via constraint~\eqref{eqn:inter-task-travel}, which ensures that when an agent performs task $j$ directly after task $i$, the tasks are scheduled with sufficient time for the agent to travel from task $i$ to $j$.
\begin{align}
	\sum_{k\in \overline{\mc M}} o^r_{kk'}     & = x^r_{k'},        &  & \forall r\in \mc R,\ \forall k' \in \mc M                                            \\
	\sum_{k'\in \mc M} o^r_{kk'} + z^r_k       & = x^r_{k},         &  & \forall r\in \mc R,\ \forall k \in \overline{\mc M}                                  \\
	\sum_{k\in \overline{\mc M}} x^r_k\, z^r_k & = 1,\;x^r_0=1      &  & \forall r\in \mc R                                                                   \\
	o^r_{kk'}\left(S_{k'} - F_k\right)         & \geq \Delta_{kk'}, &  & \forall k \in \overline{\mc M},\ \forall k' \in \mc M  \label{eqn:inter-task-travel} \\
	F_k                                        & \leq \tau,          &  & \forall k\in \mc M
\end{align}

Given the computational complexity of solving a mixed-integer problem with non-linear constraints, this approach is more applicable to instances with sparser precedence relationships and fewer robots.
Nevertheless, as solving the MINLP provides an optimal solution when solved to convergence, it serves as a benchmark for heuristic solvers.
The MINLP formulation admits a trivial solution corresponding to none of the tasks being executed.
Thus, when solved using methods such as branch-and-bound, it always maintains a set of feasible solutions, and the solver can be terminated \textit{anytime} to obtain a non-optimal feasible solution.

\subsection{Flow Model}%
\label{sec:flow-model}%
The MRTA problem described in~\scref{subsec:ps} over the directed acyclic task graph as a min-cost network flow problem.
By reinterpreting the robot coalitions assigned to tasks as flows between edges in the task graph, we compute a set of reward-maximizing flows $\{f_{ij}\}$ along the edges $(i,j) \in \mc{E}$.

\begin{definition}[Robot Flows]%
	\label{def:robot-flows}
	Let $f_{ij} \in[0, 1]$ represent the \textit{population fraction}, or proportion of the total number of agents available for the mission $N$, flowing along edge $(i, j)$.
	The robots represented by flow $f_{ij}$ are assigned to complete task $T_i$, followed by task $T_j$.
	Additionally, the total available flow at the source node $0$ is equal to 1~\eqref{eq:global-flow-constraint}, where $\Nout[i] = \{T_j\mid (i, j) \in \mc{E}\}$.

	\noindent%
	\begin{minipage}{0.2\textwidth}%
		\begin{equation}%
			\label{eq:flow-def}
			\sum_{i \in \Nin[j]} f_{ij} = \frac{C_j}{N}
		\end{equation}
	\end{minipage}%
	\hspace{1.1cm}
	\begin{minipage}{0.2\textwidth}%
		\begin{equation}%
			\label{eq:global-flow-constraint}
			\sum_{k\in \mc{N}_0^{\mathrm{out}}} f_{0k} \leq 1
		\end{equation}
	\end{minipage}
\end{definition}

The number of agents flowing out of a node cannot exceed the number of agents that flow into that node, as given by the following flow constraints:

\begin{equation}%
	\label{eq:nodewise-flow-constraint}
	\sum_{k \in \Nout[j]} f_{jk} \leq \sum_{i \in \Nin[j]} f_{ij} \quad \forall j \in \mc{M}
\end{equation}
The representation of task assignments with population fractions results in solution methods that are \textit{agnostic to the number of agents in the mission}; larger total coalition sizes do not increase computational complexity.

No explicit precedence constraints are necessary in this formulation because precedence relationships are encoded in the graph topology and reward functions. 
Note that the inequality in~\eqref{eq:nodewise-flow-constraint} allows robots to leave the coalition if necessary.

Edge capacity constraints can be optionally included if there is a limit on how many robots can traverse from one task to another.
For a pair of vertices $i$ and $j$, let $c_i^j$ be the capacity of the edge, or equivalently, the fraction of the robot team that can traverse from task~$i$ to task~$j$.
Then, the following constraints address the above requirement in the MINLP and the flow model:
\begin{equation}
  \sum_{r\in \mc R}o^r_{ij} \leq c_i^j N, \text{ and} \quad f_{ij} \leq c_i^j
\end{equation}

\subsection{The Process of Modeling a Mission}
\label{sec:mission-modeling-process}
Here, we provide an overview of the steps required to model a mission using the proposed modeling approach. 
\begin{enumerate}
    \item Define tasks: determine which elements of a mission constitute tasks 
    \item Determine graph topology: draw directed edges between tasks that have a precedence relationship 
    \item Define coalition  and influence functions: using statistical modeling or expert design, assign coalition and influence functions to each task according to their relationship
    \item Define influence aggregation functions, e.g. sum for logical ``OR'', product for logical ``AND''
    \item Define coalition influence aggregation functions, e.g. sum for ``soft'' constraint, product for ``hard'' constraint
\end{enumerate}

\section{Flow-Based Solution to the Task Graph}\label{sec:flow-solution}
In this section, we develop solution methods for the flow-based representation of the multi-robot task allocation problem defined in \scref{sec:flow-model}.
First, we describe a graph pruning algorithm that prunes the task graph to remove tasks that could violate the makespan constraint.
We then describe our primary approach to solving the flow-based task allocation problem: using an off-the-shelf non-linear programming solver.
We also introduce a greedy flow-based solution method to serve as a basis for comparison.
Finally, we develop an online flow-based solution method.
This approach leverages the relative computational simplicity of the flow-based problem formulation to apply the NLP solution approach in an \textit{iterative} manner.
This formulation requires modification of the task graph model and corresponding optimization problem at each iteration.

\subsubsection{Graph Pruning to Satisfy the Makespan Constraint}%
\label{sec:graph-pruning}
The makespan of the mission is constrained (\scref{subsec:ps}), which must be reflected in the network flow formulation.
Towards this end, we prune nodes from the task graph if robots cannot complete them within the makespan budget.
We generate a \textit{makespan graph} $\mc{G}_M = (\Mbar, \mc{E})$ with identical topology to the task graph $\mc{G}_T$.
For a path in the graph comprising a set of nodes $\mc{T}_P$ and edges $\mc{E}_P$, the duration for an agent to traverse the path is given by the sum of task durations and inter-task travel times: $d_P = \sum_{t \in \mc{T}_P}d_t + \sum_{(i,j) \in \mc{E}_P}\Delta_{ij}$.
Each node $j$ is labeled with a \textit{maximum} (worst-case) finish time $F_j$ corresponding to the longest duration path that terminates at that node.
This conservative definition is necessary because we assume that robot coalitions must be fully formed to begin a task.
From the graph, we prune any node $j$ for which $F_j > \tau$.
Therefore, any flow solution over the pruned graph will respect the makespan constraint.

\subsubsection{Rounding of Population Fraction}
\label{sec:round-solution}
The flow-based solution deals with the continuous measure of population fraction, whereas real-world task allocation requires discrete quantities of robots to compute a task schedule for each robot.
In order to translate flow solutions into realistic allocations of robots to tasks, we apply a rounding algorithm.
Iterating over the nodes in the task graph in a topologically sorted~\cite{DasguptaPV06book} manner, we compute a minimum error rounding of the flows along the outgoing edges that respect the flow conservation constraints~\eqref{eq:nodewise-flow-constraint}.
This results in a vector of integer flows that can be converted to individual robot schedules.

Since the task graph is directed and acyclic (DAG), the algorithmic complexity for both graph pruning and rounding steps is $\mathcal O(M+\vert \mathcal E\vert)$.

\subsection{Non-Linear Programming (NLP) Flow Solution}%
\label{sec:nlp-solution}
Our primary approach to solving the task allocation problem via network flow is through nonlinear programming, using an off-the-shelf solver~\cite{scipy}.
The objective is to maximize the task reward sum~\eqref{eq:cost} on the pruned graph via a set of flows $\{f_{ij} \mid (i,j) \in \mc{E}\}$ respecting the constraints~\eqref{eq:global-flow-constraint} and~\eqref{eq:nodewise-flow-constraint}.

\subsection{Greedy One-Step Lookahead Flow Solution}%
\label{sec:greedy}
Since the problem is NP-hard and computing solutions with optimal guarantees is prohibitively challenging for large problem instances, we develop a greedy one-step lookahead algorithm as a baseline for comparison.
The greedy algorithm starts from node $0$ and proceeds sequentially through the graph nodes in topological order.
Each node has a quantity of incoming flow $\fin[j]$, and $\fin[0] = 1$.
At each node $j$, the greedy algorithm allocates $\fin[j]$ into a vector of outgoing flows along outgoing edges from node $j$: $\bfout[j] = [f_{jk}, \quad \forall k \in \Nout[j]]$.
For the greedy algorithm, we enforce the conservation of flow, i.e., $\fin[j] = \lvert\bfout[j]\rvert_{\ell_1}$.
The outgoing flow $\bfout[j]$ is chosen to maximize the sum of rewards expected from outgoing neighbors: $\sum_{k \in \Nout[j]}r_k$.
This local maximum is computed by first taking the best of 50 random samples of the outgoing flow space.
We then perform gradient ascent on the reward sum $\sum_{k \in \Nout[j]}r_k$ to come to the final values for the outgoing edges of $T_j$.
This one-step lookahead is myopic but computationally simple.
The gradient ascent step requires task coalition functions and aggregation functions to be differentiable.

\subsection{Online Flow-Based Solution}\label{sec:online-solver}
The flow-based formulation described in this section leverages the graph structure of the multi-robot task allocation problem to reduce computational complexity, resulting in faster solution times.
  Thus far, in \Cref{sec:nlp-solution,sec:greedy} we developed \textit{offline} or \textit{open-loop} approaches: our flow-based solvers generate a solution at the beginning of the mission, which is then executed by the agents until completion.
	This approach can be problematic in environments with uncertainty, missions with imprecise task reward models, and dynamic environments where task parameters may change, which may significantly reduce the effectiveness of the initial plan.
  This motivates the development of a \textit{closed-loop} or \textit{online} algorithm to adapt to changes and actual task outcomes.
  We propose an online iterative algorithm (\Cref{alg:step}), built on the flow-based solver, which enables re-planning based on observed differences in the environment and robot failures.
	This \textit{online} nature of the algorithm feeds the observed rewards back into the reward model---the influence functions are modified to reflect reward observations.
	The algorithm then performs optimization on the modified reward model, changing the allocation plan accordingly.

Our flow-based algorithm is fast enough (solves problems in seconds) that it can be run repeatedly in a real-time setting, even on large problem sizes.
Therefore, we can run the solver whenever task reward feedback is received without incurring any significant delays.

We now introduce an iterative formulation of our flow-based NLP solver that performs a step after each task is completed, modifying and re-solving the task graph problem to incorporate new information observed.
This formulation involves several major modifications to the modeling and solution approach:
\renewcommand{\labelenumi}{\alph{enumi})}
\begin{enumerate}
	\item We modify the graph topology to reflect tasks that are completed and add new source nodes.
	\item We modify the reward model defined over the task graph, to reflect rewards observed from completed tasks.
	\item We modify the NLP Solver constraints to reflect the above changes in topology and reward model.
	\item We implement a redundancy scheme in which solutions from prior steps are projected onto the updated graph and checked against the new solution. 
\end{enumerate}

\subsubsection{Graph Topology Modifications for Online Solver}%
\label{sec:online-topology-modifications}
Each step is triggered by a coalition of agents completing a task.
We make several modifications to the graph topology in order to reflect this event.
First, the completed task is removed from the task graph, along with the associated incoming and outgoing edges.
If any preceding tasks remain with incoming edges to the completed task, these nodes are removed as well, and return a zero reward---to complete them after the subsequent task would violate the precedence relationship.

Second, a new artificial source node is created that gathers all free agents, i.e. the agents that have just completed a task or are in an idle state.
The new artificial source node is connected with new edges to (i) all tasks without any incoming edges and to (ii) all tasks that were previously connected to the task that was just completed.

Third, we modify the graph to ensure that robot coalitions working on in-progress tasks remain assigned to those tasks.
This preserves the continuity of the task plan and ensures that the model accounts correctly for the time remaining to complete the in-progress tasks.
To accomplish this, we add a new artificial source node to each in-progress task.
The source node connects only to its associated task and has as many robots as the coalition currently working on the in-progress task.
This, along with solver constraints laid out in \scref{sec:solver-constraint-mods}, ensures that the same coalition is assigned to the in-progress tasks.

The addition of new source nodes and removal of completed task nodes creates a distinct task graph $\mc{G}_{T,i} = (\Mbar_i, \mc{E}_i)$ and associated reward mode for each iteration $i$.

\begin{figure}
	\centering
	\includegraphics[trim={0.0cm 0.0cm 0.0cm 0.0cm},clip,width=\linewidth]{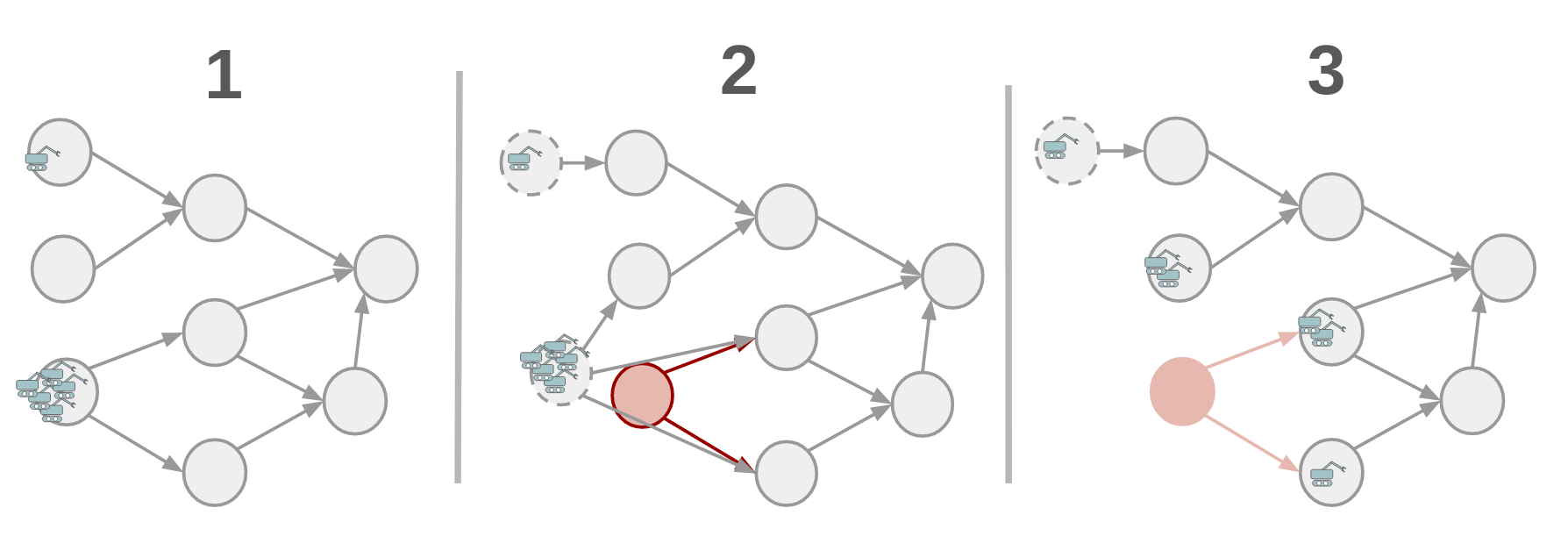}
	\caption{\label{fig:online-topology}
		In Step 1, the team of five agents complete their task.
		In Step 2, the graph is modified to remove the completed task node (indicated in red) and add an artificial source node at the robots' new location.
		Meanwhile, the single-agent coalition's task is in progress, so the agent is placed on another artificial source node.
		In Step 3, the agents are reassigned, travel to their new tasks, and start work.
		Note that the configuration in Step 3 is not accessible from Step 1 without the online iterative solver due to the flow constraints.
	}
\end{figure}

\fgref{fig:online-topology} shows an example of the graph topology modification that occurs during an iteration of the online solver.
Note that the configuration reached in pane 3 is unachievable from the configuration at pane 1 while respecting the flow constraints, yet no precedence constraints were violated.
This demonstrates the ability of the iterative formulation of the flow-based solver to jump between branches of the task graph, while ensuring all precedence constraints are enforced.

\subsubsection{Reward Model Modifications}
\label{sec:online-reward-model-mods}
At each step, we make several modifications to the reward model to ensure that the newly observed reward $r_i$ is incorporated.
First, we modify the reward models of all outgoing neighbors $j \in \mc{N}_{i,out}$ of the newly completed task.
For a given neighbor $j$, if the task has no remaining incoming neighbors, we calculate the influence value from the completed task $\delta_{ij}(r_i)$, and aggregate it with the coalition to produce a new reward model: $r'_j = \delta_{ij}(r_i)\ddagger_j \rho_j(C_j)$.
If incoming edges remain to task $j$, then the reward~$r_i$ is integrated into the influence function as a constant.
\begin{remark}
The impacts of failed or partially incomplete tasks are captured via this modification. If task $p$ fails, it returns $r_p=0$, which is incorporated into an updated reward model, thereby, capturing the failure's impact on downstream tasks.
\end{remark}
We also shorten the task duration of in-progress tasks to reflect the time that has passed since each task commenced, and update the inter-task travel times of new edges to reflect the distances between prior agent locations and the newly connected tasks.

\subsubsection{Flow-based NLP Solver Constraint Modifications}%
\label{sec:solver-constraint-mods}
The addition of new artificial source nodes as described in~\scref{sec:online-topology-modifications} necessitates several modifications to the flow-based NLP solver constraints from~\eqref{eq:global-flow-constraint} and~\eqref{eq:nodewise-flow-constraint}.
To make these modifications, we divide the graph nodes into three mutually exclusive sets: $\mc{T}_0 = \{T_0\}$ is the source node set.
It includes only node zero, the source node from which free agents are allocated.
$\mc{T}_I = \{T_{n_s},\ldots,T_M \}$ is the set of \textit{internal} nodes that are neither source nodes nor sink nodes, where $n_s$ indicates the total number of artificial source nodes including node~0.
$\mc{T}_S = \{ T_1,\ldots,T_{n_s} \}$ is the set of artificial source nodes corresponding to \textit{in-progress tasks}, with cardinality $n_s -1$, i.e., the set is empty if there are no in-progress tasks.
The set of in-progress task source nodes $\mc{T}_S$ is accompanied by a set of source node capacities $C_S = \{\frac{C_i}{N} \mid i \in {1,\ldots,n_s}\}$, where $C_i$ is the size of the coalition currently working on in-progress task $i$, and $N$ is the total number of robots in the mission.

First, we constrain the flow leaving the source node to be less than or equal to the coalition fraction of free agents (those not working on any in-progress tasks).

\begin{equation}%
	\label{eq:online-global-flow-constraint}
	\sum_{k\in \mc{N}_0^{\mathrm{out}}} f_{0k} \leq 1 - \sum_{j=1}^{n_s}\frac{C_j}{N}
\end{equation}

For each source node affiliated with an in-progress task, we constrain the total outflow from the node to be exactly equal to $C_j/N$, the quantity of flow currently assigned to the in-progress node $j$.
Note that the set $\mc{N}_i^{out}$ will always contain only one member: the affiliated in-progress task itself.

\begin{equation}%
	\label{eq:online-inprogress-flow-constraint}
	\sum_{k\in \mc{N}_i^{\mathrm{out}}} f_{jk} = \frac{C_j}{N} \quad \forall j \mid T_j \in \mc{T}_S
\end{equation}

We modify the node-wise flow constraint to apply only to the set of internal nodes $\mc{T}_I$, not artificial source nodes.
\begin{equation}%
	\label{eq:online-nodewise-flow-constraint}
	\sum_{k \in \Nout{j}} f_{jk} \leq \sum_{i \in \Nin{j}} f_{ij} \quad \forall j \mid T_j \in \mc{T}_I
\end{equation}

Under these constraints, the solutions produced by the flow-based solver in each iteration will be compatible with the current state of the robots and in-progress tasks.

\subsubsection{Projecting Solutions onto New Graphs}\label{sec:projecting-solutions}%
The nonlinear solver used in the flow-based NLP approach occasionally falls into local minima and generates poor solutions.
To prevent this, we implement a redundancy scheme: during each step of the online solver, we check the new solution against solutions from prior steps.
We project the prior solution on the current task graph, and if the generated reward is higher than the current solution, the projected solution is selected.
We develop \Cref{alg:check-and-update}, which projects a flow-based solution from a prior task graph onto the edges of the current graph.

The projection algorithm iterates through the edges of the current task graph $\mc{E}$ and assigns flow values according to a previous solution $\mb{f}' \in \mathbb{R}^{\lvert\mc{E}'\rvert}$ from a previous iteration and its associated task graph $\mc{G}'_{T} = (\Mbar', \mc{E}')$.
This generates a solution $\mb{f}$ in the current solution domain $\mb{f} \in \mathbb{R}^{\lvert\mc{E}\rvert}$.

First, the algorithm considers the edges leaving in-progress task source nodes $\mc{T}_S$ (defined in the previous section).
For an in-progress task node $T_q \in \mc{T}_S$, the algorithm assigns flow to its associated edge $(p,q)$ according to the flow through $T_q$ in the prior solution $\mb{f}'$: $f_{pq} = \sum_{j \in \Nin[q]}f'_{jq}$, where the neighborhood $\Nin[q]$ is calculated according to graph $\mc{G}_T'$.

Second, the edges leaving the free agent source node $T_0$ are assigned flow values.
For each node $T_q$ in the outgoing neighborhood of the source node in the current graph $\mc{N}_0^{out}$, we assign edge $(0,q)$ the total flow through that node in the previous solution $\mb{f}'$: $f_{0q} = \sum_{j \in \mc{N}_q^{in}}f'_{jq}$.

Third, all remaining edges that have not yet been assigned flow values in the new graph are ``internal'' edges that are not connected to source nodes.
These edges are not new to the task graph, and therefore, they exist in the previous edge set $\mc{E}'$.
For each of these edges $(p,q)$, the algorithm assigns the flow value from the previous graph $f'_{pq}$ to the corresponding edge in the new graph $f_{pq}$.

The algorithmic complexity of the first three modification steps is $\mc  O(M+\vert \mc E \vert)$, as they require a simple traversal of the directed and acyclic task graph. The final redundancy step has $\mc O(\vert \mc E \vert)$ complexity for each prior solution, and since there are a maximum of $M$ prior solutions, the overall complexity is $\mc O(M \vert \mc E\vert)$. The complexity of the algorithm also depends on the external NLP solver; let's denote it by the function $f_{\text{NLP}}$, which depends on the current task graph and mission model. The overall complexity of the online solver is then $\mc O(M \vert \mc E\vert f_{\text{NLP}})$.

\begin{algorithm}[htpb]%
	\small
	\newcommand{\tcand}{\mathrm{cand}}
	\Input{Newly completed task $T_{nc}$, completed task reward $r_c$, set of completed tasks $\mc{T_C}$, set of rewards observed $\mb{r}_{obs}$ in-progress task set $\mc{T}_{IP}$, sets of source nodes $\mc{T}_0$ and $\mc{T}_S$,  in-progress coalition set $C_{IP}$, set of in-progress task times elapsed $t_{IP}$, current time $time$, set of prior solutions and graphs $\{(\mb{f}_k, \mc{G}_k) \; | \; k \in \{0,1,...,\;i-1\}\}$}
	\Output{Flow solution $\mb{f}_i$, updated task graph $\mc{G}_i = (\Mbar_i,\mc{E}_i)$}
	\tcp{\textbf{Modify graph topology: \scref{sec:online-topology-modifications}}}
	$\mc{G}_i = (\Mbar_{i},\mc{E}_{i}) \gets (\Mbar_{i-1} - T_{nc},\mc{E}_{i-1})$\\
	\For{$(p,q) \in \mc{E}_i$}{
		\uIf{$p \in \mc{T}_C$ \KwOr $p \in \mc{T}_0 + \mc{T}_S$ \KwOr $q\in \mc{T}_C$}{
			$\mc{E}_i \gets \mc{E}_i - (p,q)$
		}
		\uIf{\KwNot $q \in \mc{T}_C$}{
			$\mc{T}_C \gets \mc{T}_C \cup \{q\} $\\
			$\mb{r}_{obs}[q] \gets 0.0$
		}
	}
	\tcp{Rename nodes and edges by adding $\lvert\mc{T}_{IP}\rvert$ to all node indices}
	source\_ct $\gets 1$\\
	\For(\tcp*[f]{in-progress tasks}){$t \in \mc{T}_C$}{
		$\mc{E}_i \gets \mc{E}_i \cup \{($source\_ct$,t)\}$
	}
	\ForEach{$\{t \in \Mbar_i\text{ s.t. } \lvert\Nin[t]\rvert=0\}$}{
		$\mc{E}_i \gets \mc{E}_i \cup \{(0,t)\}$
	}
	\For{$t \in \Nout[T_{nc}]$}{
		$\mc{A}_t \gets $ add\_ghost\_reward$(r_c)$ \tcp{\scref{sec:online-reward-model-mods}}
	}
	\tcp{Update in-progress task durations}
	\For{$t \in \mc{T}_{IP}$}{
	$d_{t,i} \gets d_{t,i-1}-t_{IP}[t]$
	}

	\tcp{\textbf{Perform graph pruning (\scref{sec:graph-pruning})}}
	$\mc{G}_i = (\Mbar_{i},\mc{E}_{i}) \gets $ prune\_graph$(\mc{G}_i)$ \\
	\tcp{\textbf{Create and solve optimization problem (\scref{sec:flow-solution})}}
	NLP $\gets$ non-linear program object \\ NLP.add\_objective(reward\_model$_i$) \\ NLP.add\_constraint($\sum_{k \in \Nout[0]}f_{0k} \leq 1 - \sum_{j=1}^{|\mc{T}_S|} \frac{C_j}{N}$)\\ \For{$t \in \mc{T}_S$}{ NLP.add\_constraint($\sum_{j \in \Nout[t]}f_{tj} = \frac{C_j}{N}$) } \For{$t \in \mc{T}_I$}{ NLP.add\_constraint($\sum_{k \in \Nout[t]}f_{tk} \leq \sum_{j \in \Nout[t]}f_{jt}$) } $\mb{f}_i' \gets $ NLP.solve()\\ $\mb{f}_i \gets $ round\_solution($\mb{f}_i'$) \tcp{See \scref{sec:round-solution}} \tcp{\textbf{Check and update solution against prior solutions (\scref{sec:projecting-solutions} -- see \Cref{alg:check-and-update})}} $\mb{f}_i^* \gets $ check\_and\_update\_solution(($\mb{f}_i$, $\mc{G}_i$), $(\mb{f}_k, \mc{G}_k) \forall k \in \{0.
		.i-1\}$, $\mc{T}_{IP}$) \\
	\Return $\mb{f}_i^*$, $\mc{G}_i$
	\caption{Online flow-based solver step}%
	\label{alg:step}
\end{algorithm}

\begin{algorithm}[htpb]%
	\small
	\newcommand{\tcand}{\mathrm{cand}}
	\Input{Current solution and corresponding graph ($\mb{f}_i$, $\mc{G}_i$), set of prior solutions and graphs $\{(\mb{f}_k, \mc{G}_k) \mid k \in \{0,1,\ldots,i-1\}\}$, in-progress task set $\mc{T}_{IP}$}
	\Output{Best flow solution $\mb{f}_i$}
	$\mc{R}_{\tcand} \gets \{\}$ \tcp{Set of candidate rewards}
	$\mc{F}_{\tcand} \gets \{\}$ \tcp{Set of candidate solutions}
	\For{$(\mb{f}', \mc{G}') \in \{(\mb{f}_k, \mc{G}_k) \mid k \in \{0,1,\ldots,i-1\}\}$}{
	$\mb{f}^{\tcand} \gets \mb{f}'$ \tcp{Candidate projections}
	$\mc{E}_u \gets \{\}$ \tcp{Set of updated edges}
	\For{$q \in \mc{T}_{IP}$}{
	$p \gets n \in \mc{T}'_S \; \mid \; n \in \Nin[q]$ \tcp{source node for task $q$ in graph $\mc{G}'$}
	$f_{pq}^{\tcand} \gets \sum_{n \in \Nin[q]}f_{nq}'$\\
	$\mc{E}_u \gets \mc{E}_u \cup \{(p,q)\}$
	}
	\For{$p \in \Nout[0]$}{
	$f_{0p}^{\tcand} \gets \sum_{j \in \Nin[p]}f_{jp}'$ \\
	$\mc{E}_u \gets \mc{E}_u \cup \{(0,p)\}$
	}
	\For{$(p,q) \in \mc{E}_i - \mc{E}_u$}{
		$f_{pq}^{\tcand} \gets f_{pq}'$
	}
	$\mc{R}_{\tcand} \gets \mc{R}_{\tcand} \cup \{$reward\_model$(\mb{f}^{\tcand})\}$
	}
	\Return $\mc{F}_{\tcand}[\operatorname{argmax}(\mc{R}_{\tcand})]$
	\caption{check\_and\_update\_solution\\(\scref{sec:projecting-solutions})}%
	\label{alg:check-and-update}
\end{algorithm}

\subsection{Offline vs. Online Solver} The online formulation has two primary benefits.
First, when unexpected observations occur, the online solver incorporates the new information into the reward model and generates a new plan that reacts best to the disturbance.
This ability is beneficial whenever the reward model is inaccurate or disturbances occur, but especially when missions are sensitive to particular tasks.
For example, a mission might include ``explore an area to find a path through,'' followed by a set of tasks on the other side of the unexplored region.
If the exploration task fails and no route is found, every subsequent task will yield zero reward.
The online solver is able to reallocate resources given this information and pull agents to tasks 
that still might yield reward.

Second, the iterative application of the online solver broadens the solution domain.
Consider the sample task graph shown in~\fgref{fig:online-topology}.
For an individual application of the flow-based solver, solutions must obey the flow constraints, and therefore can only move along edges present in the task graph.
This constraint ensures precedence constraints are fulfilled and decreases the computational complexity of the problem, but restricts the problem domain.
However, in repeated applications of the flow-based solver, we modify the task graph, \textit{adding new edges} as robots finish tasks and are located on new artificial source nodes, which are then connected to the rest of the graph.
The method of connecting these artificial source nodes described in \scref{sec:online-topology-modifications} ensures that precedence constraints are not violated.
But the addition of these new edges still allows robots to \textit{traverse between branches of the task graph}, even when those branches were not initially connected by a precedence relationship (edge).
This is demonstrated in~\fgref{fig:online-topology}, where robots end in pane 3 in a configuration unreachable from the initial state in pane 1: they have crossed from a lower branch to a higher branch when these branches were not connected in pane 1.
This capability to add edges and traverse branches while still fulfilling the precedence relationships increases the solution domain.
There are possible allocations in the online solver formulation that would violate constraints in the offline solver formulation.

The price of these benefits is computational complexity.
The online solver applies the flow-based solver at every step---every time a task is completed.
In the worst case, when every task in a mission is completed, the flow-based solver is run $n-1$ times, where $n$ is the number of total tasks in the mission.
There are two ameliorating factors to this computational complexity increase that bear consideration: (1) every time a task is completed, the problem shrinks, until it is trivial when only a few tasks remain, and (2) the online solver is applied throughout the duration of a mission, rather than just at mission initialization -- computation time must be contextualized within the individual task durations, inter-task travel times, and overall mission time.

\section{Simulation Studies}\label{sec:simulation-studies}
We develop a testbed that randomly generates task graph missions for the evaluation of our solution approaches while varying problem parameters, including mission size, number of agents, makespan, and random error characteristics.
In Experiments 1--3, we evaluate the solvers in an \textit{offline} setting: we compare the MINLP solver with the greedy and NLP flow-based solvers without error in the reward model or feedback.
In Experiments 4--6, we compare the offline and online formulations of the flow-based solver: we evaluate their performance compared to near-optimal results from the MINLP solver, and compare them to one another when zero error is present.
We then develop two models for uncertainty in the system---stochastic task failures and stochastic model error---and compare the online and offline flow-based solvers with several baselines in Experiments 7 and 8.

\subsection{Offline Task Allocation Experiments}\label{sec:offline-experiments}%
In our offline set of experiments, we evaluate the flow-based, MINLP, and greedy algorithms on randomly generated task graphs.
We assume that the rewards predicted by the task graph reward model are accurate.
We neglect inter-task travel time for Experiments~1--3 only and incorporate it for all other experiments in this article.
We generate solutions with all three algorithms at planning time and then execute the solutions open-loop to measure performance.
Experiments 1--8 were run on a desktop computer with an 8-core Intel Core i7-9700 CPU.
Our implementation uses the PySCIPOpt optimization suite~\cite{MaherMiltenbergerPedrosoetal.2016, BestuzhevaEtal2021OO} as the solver for the MINLP approach and SciPy optimize~\cite{scipy} for the flow-based NLP approach.

\textbf{Random Graph Generator Platform and Experiments:}
We developed a random graph generator to generate graphs with random topologies and reward functions, allowing us to explore the mission specification space broadly. We specify a target number of tasks, and randomly generate coalition and influence functions. We sample from polynomial, sub-linear, and sigmoid functions (a smoothed approximation of a step function) with randomized parameters.
We randomly sample one of the sum and product functions as the aggregation function, $\ddag$, which combines the coalition and influence function outputs. For these experiments, we use the sum operation as our influence aggregation function $\alpha_j$ for each task $T_j$.

\subsubsection{Experiment 1: Number of Tasks}

In this experiment, we evaluate the impact of the number of tasks in the mission on the relative performance of the three solution methods.
For each number of tasks, we conduct ten trials consisting of unique randomly generated task graphs.
For this experiment, we use 4 agents, a makespan constraint $\tau = 0.6\sum_{j \in \mathcal{M}}d_j$, and limit computation time to 10~minutes per trial.

\begin{figure}[tbp]
  \centering
  \includegraphics[width=\columnwidth]{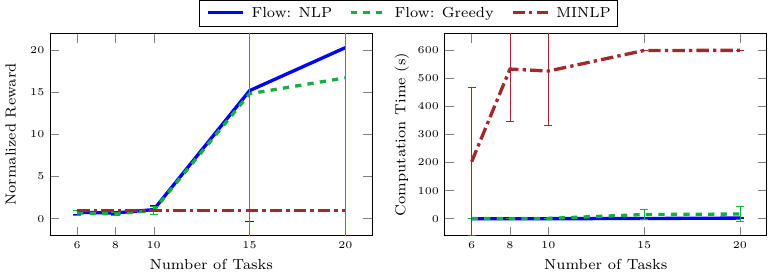}
  \caption{\label{fig:n_tasks_partial} Experiment 1 (partial domain) tests planner performance as mission size (number of tasks) is increased up to 20 tasks such that the MINLP approach, limited by 10~minutes of computation time, generates non-trivial solutions.
      We show the rewards of the flow-based NLP and greedy approaches normalized by the MINLP reward.
  As mission size grows, the flow-based approaches outperform the MINLP approach, which fails to converge to an optimal solution.
The computation time of the flow-based approaches is significantly less than that of the MINLP approach.}
\end{figure}

\begin{figure}[htbp]
  \centering
  \includegraphics[width=\columnwidth]{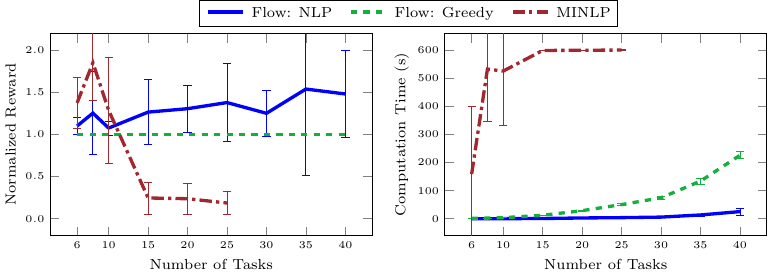}
  \caption{\label{fig:n_tasks_full} Experiment 1 (full domain) tests planner performance as mission size (number of tasks) is increased up to 40 tasks.
  The rewards of solutions computed by the MINLP and flow-based NLP approaches are shown normalized by that of the flow-based greedy approach.
The NLP approach outperforms the greedy approach by 30\% on average and is 10 times faster.}
\end{figure}

\textbf{Partial Domain:} In~\fgref{fig:n_tasks_partial}, we examine the portion of the domain in which the MINLP solution consistently computes a non-trivial feasible solution: $\leq 20$ tasks.
We graph the mean of rewards normalized by the MINLP reward for each trial, e.g., if the reward from the flow NLP solution is 20\% higher than the reward from the MINLP solution, we report $1.2$.

\textbf{Full Domain: } In~\fgref{fig:n_tasks_full}, we show the full domain we tested, normalize the results by the \emph{greedy} flow solution, and graph the MINLP results only when feasible and non-trivial. On average, over all trials in the full domain, the flow-based NLP approach outperforms the greedy approach by 30\% and takes 10\% the time of the greedy approach to compute a solution. 

\subsubsection{Experiment 2: Number of Agents}
\begin{figure}[tbp]
  \centering 
  \includegraphics[width=\columnwidth]{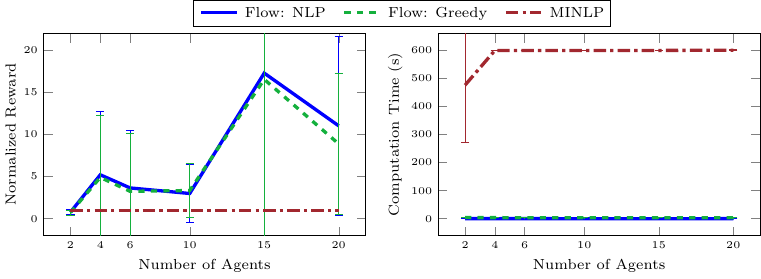}
  \caption{\label{fig:n_agents_exp} Experiment 2 demonstrates the impact of the number of agents in the coalition on the performance of the three solution methods. We plot the reward (left) and computation time (right).
    The rewards are normalized by the MINLP solution.
  As the number of agents increases, the MINLP problem size grows, whereas the flow-based solutions only get more accurate, due to decreased rounding error.}
\end{figure}

Experiment 2 evaluates the relative performance of the solution methods as the number of agents varies. We test with total agent populations of \{2,4,6,10,15,20\}, and conduct 10 trials with unique randomly generated task graphs for each value.
For this experiment, we use 10 tasks, a makespan constraint $\tau = 0.6\sum_{j \in \mathcal{M}}d_j$, and limit computation time to 10 minutes per trial.
In~\fgref{fig:n_agents_exp}, we plot the rewards, normalized by the MINLP reward, and the computation time for each method.

\subsubsection{Experiment 3: Makespan and Number of Tasks}
\begin{figure}
  \centering 
  \includegraphics[trim={0.2cm 0.0cm 0.2cm 0.0cm},clip,width=\linewidth]{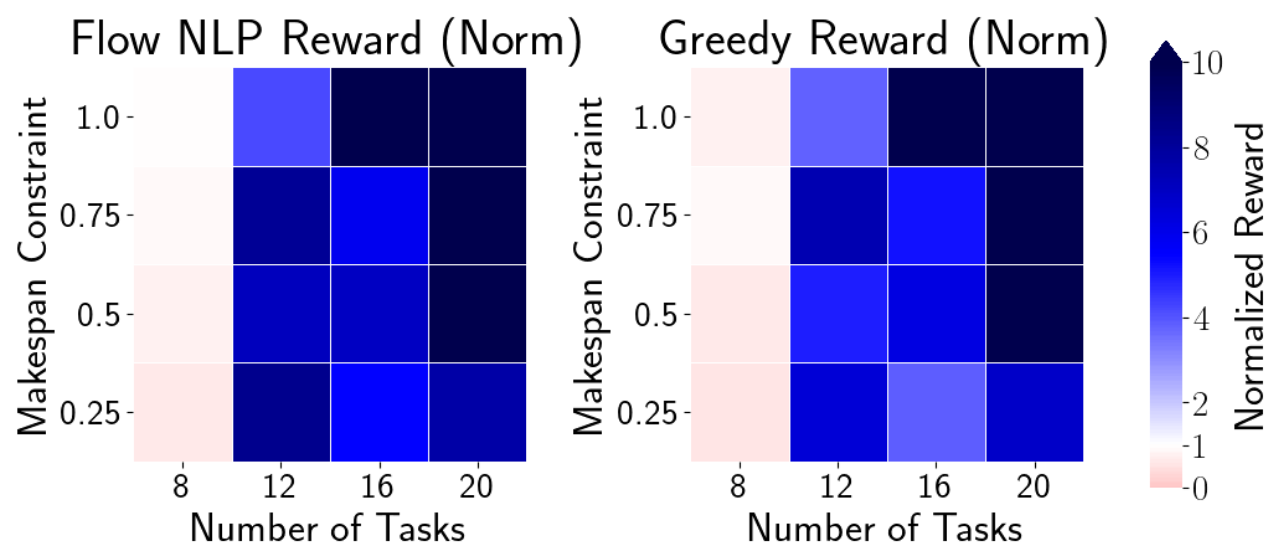}
  \caption{\label{fig:makespan_exp} The reward accrued by the NLP flow approach (left) and the greedy flow approach (right) for varying makespan constraint and the number of tasks.
  The rewards are normalized by the MINLP reward. The flow-based approaches perform worse than the MINLP on small problems with a tight makespan constraint, but outperform on problems larger than 8 tasks.
The MINLP approach fails to compute non-trivial solutions to larger problems than those shown here.}
\end{figure}

In Experiment 3, we examine the impact of both the makespan constraint and the number of tasks on the relative performance of the solution methods. In this experiment, we examine \textit{only the portion of the domain in which the MINLP approach computes non-trivial solutions.} We formulate the makespan constraint for randomly generated graphs as a proportion of the total duration of all tasks, $\mu \sum_{j \in \mc{M}}d_j$. We evaluate $\mu=\{0.25,0.5,0.75,1.0\}$ and $M=\{8,12,16,20\}$ tasks and conduct 10 trials at each pair of values. For this experiment, we use 4 agents and limit computation time to 10 minutes per trial. We show the NLP and greedy flow solution results normalized by the MINLP results in~\fgref{fig:makespan_exp}.

\textbf{Offline Experiment Discussion:}
These experiments show that the flow-based algorithms perform better and faster than the MINLP approach when the problem is moderately large ($>12$ tasks or $>4$ agents). The MINLP fails to converge for large problems ($>$20 tasks, $>$20 agents), whereas the flow-based algorithms perform well beyond this domain. 

Furthermore, these experiments show that the computation time of the MINLP approach is orders of magnitude larger than that of the flow-based algorithms. This suggests that the flow-based solution would be appropriate for real-time or dynamic settings, or for use in iterative allocation methods, whereas previous solutions would not have been fast enough to use.

These results suggest that in this problem regime, restricting robot assignment plans to be along paths of precedence relationships, i.e., to obey the flow constraints of the graph, is a meaningful and significant simplification of the problem. 

\subsection{Online Task Allocation Experiments}
\label{sec:online-experiments-simplified}

In these experiments, we characterize the performance of our two primary solvers, the offline flow-based (Offline) and online flow-based (Online), under several experimental error conditions. First we evaluate the relative performance of the solvers when the task graph models are \textit{perfectly accurate}, i.e. they have zero error. We measure the empirical optimality of the Offline and Online algorithms by comparing to near-optimal solutions in a small number of trials in which the MINLP solver converged to near-optimality (given 12 hours). We also evaluate relative performance of the Offline and Online solvers while varying team size and mission size. 

To characterize the response to error, we also evaluate the impact of stochastic task failure and of stochastic model error on relative solver performance while varying mission size. We compare our two primary solver implementations against baseline solvers that experience no failure (NoFail-Off and NoFail-On) and to clairvoyant solvers that have access to the ground truth reward model (Clair-Off and Clair-On). We also measure the computation times of each solver as mission size changes. This suite of experiments investigates how the online formulation presented in this article changes solver performance in the face of uncertainty. It builds on the experiments in \scref{sec:offline-experiments}, as the Offline solver here is the NLP Flow-based solver developed in that section.

\subsubsection{Experiment 4: Comparison to Near-Optimal Solutions from the MINLP Solver}
We perform a limited evaluation of the online and offline flow-based solvers against the Mixed Integer Non-Linear Programming (MINLP) solver described in \scref{sec:minlp}, which yields globally optimal solutions when it converges. Due to its long solve times and associated poor performance on larger missions (see \scref{sec:offline-experiments}), this method is not scalable or applicable in an uncertain environment---re-planning around unexpected observations would take hours. However, if we assume that the given mission model is perfectly accurate, the MINLP solution gives us an optimal basis of comparison that we use to estimate the optimality of our flow-based solvers. 

As opposed to the prior experiments in \scref{sec:offline-experiments} that compare against the MINLP solver, in these results we include only trials in which the MINLP solver converges within a certain threshold.
We generate several hundred random missions with 10, 12, and 15 tasks and four agents, and apply the online flow-based (Online), offline flow-based (Offline), and MINLP (also offline) solvers to each random mission.
We run the MINLP solver until convergence with a timeout of 12 hours, and present only those trials (approximately 17\%) in which the solver converges to within a threshold.
In \Cref{table:minlp-results} we list the results from 458 total trials (the quantity possible due to high computational load on shared computing resources).
For each number of tasks, we provide average data for those trials in which the MINLP solver converged within a threshold of 1\%, 5\%, 10\%, and 50\% where appropriate. The MINLP solver uses a branch-and-bound method that generates a primal and dual bound for the solution. The convergence percentage is defined as $\frac{dual-primal}{primal}$. For example, if the convergence threshold is 1\%, we know that the true optimal value of the mission is no greater than 1.01 times the solution value generated by the MINLP solver. For each convergence threshold, we list the number of trials that converged to within this threshold, and the rewards accrued by the offline and online solvers, normalized by the MINLP solution for each trial. Performance of the online solver is consistently within 5--15\% of the MINLP solution for all mission sizes, whereas the offline solution decreases in solution quality from 70\% to less than 40\%  as mission size increases.
\begin{table}
  \renewcommand{\arraystretch}{1.3}
  \centering
  \caption{Comparing online and offline solvers to the MINLP solution approach with a 12 hour timeout.}%
  \label{table:minlp-results}
  \begin{tabular}{@{}rrrrrr@{}}
    \multicolumn{1}{c}{\multirow{2}{*}{\# Tasks}} & \multicolumn{1}{c}{\multirow{2}{*}{\# Trials}} & \multicolumn{1}{c}{\multirow{2}{*}{\begin{tabular}[c]{@{}c@{}}Convergence\\ Threshold\end{tabular}}}& \multicolumn{1}{c}{\multirow{2}{*}{\begin{tabular}[c]{@{}c@{}}\# Trials\\ Included\end{tabular}}}& \multicolumn{2}{c}{Normalized Rewards}                   \\ \cmidrule(l){5-6} 
    \multicolumn{1}{c}{}                          & \multicolumn{1}{c}{}                           & \multicolumn{1}{c}{}                                       & \multicolumn{1}{c}{}                                    & \multicolumn{1}{r}{Offline} & \multicolumn{1}{r}{Online} \\

    \toprule
    \multirow{3}{*}{10 Tasks} & \multirow{3}{*}{152} & 1\% & 19 & 74.6\% &\ 93.6\% \\
                              & & 5\% &  33 &  70.2\%  & 88.7\% \\ 
                              & & 10\% & 43 & 69.6\% & 95.6\% \\
                              \midrule
    \multirow{3}{*}{12 Tasks} & \multirow{3}{*}{129} & 1\% & 0 & ---&--- \\
                              & & 5\% &  8 &  53.3\%  & 92.7\% \\ 
                              & & 10\% & 17 & 53.6\% & 92.1\% \\ 
                              \midrule
    \multirow{4}{*}{15 Tasks} & \multirow{4}{*}{177} & 1\% &  0& --- &---  \\
                              & & 5\% & 0  & ---  & --- \\ 
                              & & 10\% & 1 & 17.9\% & 85.1\% \\ 
                              & & 50\% & 18 & 38.3\% & 91.1\% \\ 
                              \bottomrule
  \end{tabular}
\end{table}

\subsubsection{Experiment 5: Perfect Model}
We examine the performance of the offline and online flow-based solvers under \textit{zero experimental error}: the reward models that the solvers use are perfectly accurate. In this condition, the only differences between the two solvers are the expanded solution domain of the online solver, and the increased computation of the online solver -- because the online solver repeatedly applies the flow-based solver to increasingly small sub-graphs as tasks are completed and removed, stochastic aspects of optimization may result in higher performance.

\begin{figure}
	\centering 
	\includegraphics[trim={0.0cm 0.0cm 0.0cm 0.0cm},clip,width=\linewidth]{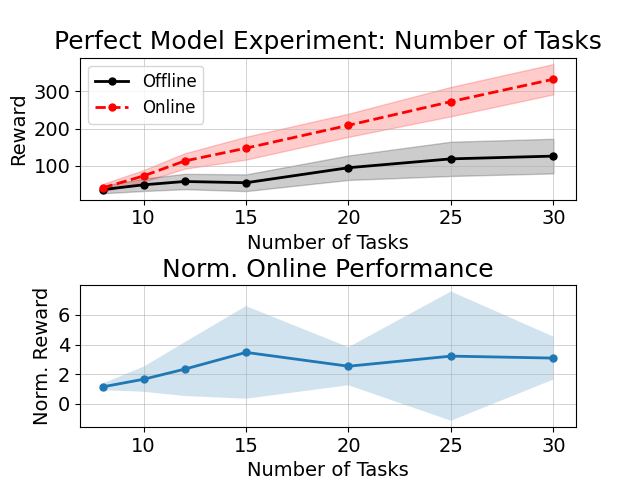}
	\caption{\label{fig:zero-err-n-tasks} We show the performance of the offline and online flow-based solvers as the mission size is varied. We test mission sizes $\{8, 10, 12, 15, 20, 25, 30\}$. The standard deviation is shaded above and below each data point. In the top graph, we plot the absolute reward means of both controllers. In the bottom graph, we show the performance of the online solver normalized by the performance of the offline solver on each trial, i.e., by what factor does the online solver outperform the offline solver.}
\end{figure}

We perform two experiments: in Experiment 5A, shown in \fgref{fig:zero-err-n-tasks} we examine the relative performance of the two solvers as the number of tasks in the mission varies. We test 100 randomly generated missions for each mission size in $\{8, 10, 12, 15, 20, 25, 30\}$, and show the mean and standard deviation over the 100 trials for each controller. We also show the performance of the online solver normalized by the performance of the offline solver for each trial. These results show that the online solver outperforms the offline solver by 50\% for small missions (8 tasks), increasing to 200\%+ as mission size increases to 15 tasks and beyond.

\begin{figure}
	\centering 
	\includegraphics[trim={0.0cm 0.0cm 0.0cm 0.0cm},clip,width=\linewidth]{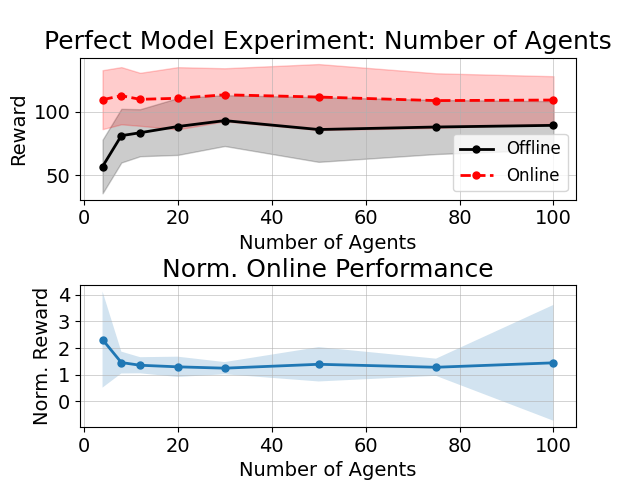}
	\caption{\label{fig:zero-err-n-agents} This experiment compares the two solvers as the number of agents in the mission is varied. In the top graph, each data point represents the average of 100 trials, and the shading represents the standard deviation. The bottom graph shows the same data, but normalizes the online solver's performance by the offline solver for each trial. This shows a decrease from outperforming the offline solver by 250\% with four agents, to outperforming by~125\% in the 20--100 agent range.}
\end{figure}

In Experiment 5B, we examine the relative performance of the two solvers as the number of agents in the mission is varied. Because the flow-based solver represents agents as a coalition fraction, the optimization problems are identical across the range of team sizes. However, due to the post-processing rounding step that translates coalition fraction into discrete agent assignments, performance varies as team size is changed. We run 100 randomly generated missions, each with 12 tasks, for each number of agents in $\{4,8,12,20,30,50,75,100\}$, and report average results for each solver. The results in \fgref{fig:zero-err-n-agents} show that the online solver consistently outperforms the offline solver. The online solver's performance is relatively constant as the team size changes, whereas the offline solver performs significantly worse at low team sizes. This demonstrates that the benefit of re-planning is accentuated at lower team sizes---the increased granularity of team assignments means that the offline solver cannot split teams among branches as effectively, resulting in tasks with suboptimal assignments. The online solver, however, has access to nearly the full solution domain, and can reassign agents across branches, thereby improving performance significantly.

These results show that even under the perfect model assumption, the online solver performs 25\% to 200\% better than the offline solver, with the largest effect seen in missions with large numbers of tasks ($\geq 15)$ and small numbers of agents ($\leq 12$). 

\subsubsection{Experiment 6: Solver Computation Time}
\begin{figure}[tbp]
  \centering 
  \includegraphics[trim={0.0cm 0.0cm 0.0cm 0.0cm},clip,width=\linewidth]{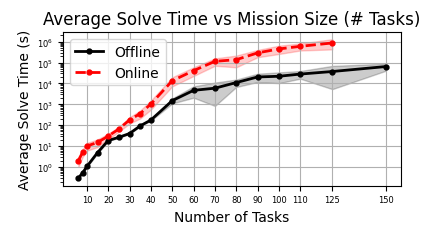}
  \caption{\label{fig:solve-times} This graph shows the average time to solve a mission with both the online and offline solvers as the mission size (number of tasks) grows. 100 randomly generated missions were averaged for each mission size up to 50 tasks. For 50-110 tasks, 10 trials were averaged, and for 125+ tasks 5 trials were averaged. Note the log scale.}
\end{figure}

In this experiment, we investigate the computation time required to generate a mission solution with the online and offline solvers as we increase the mission size (number of tasks). The results are shown in \fgref{fig:solve-times}---we generate 100 random missions for each number of tasks up to 50, 10 missions for 50-110 tasks, and 5 missions for 125+ tasks. We report the average time required by both the offline and online solvers to generate solutions for these missions. For both solvers, there appears to be an exponential relationship between mission size and computation time, although this exponential trend is significantly higher for the online solver. This is because the online solver performs a modified run of the offline solver every time a task is completed---$O(M)$ times, where $M$ is the number of tasks. 
\begin{remark}
Though long solution times limit the method's applicability with more than 100 task-missions, we note that the online solver's operation time is extended over an entire mission. The relevant metric for assessing the feasibility of applying the online solver is the ratio of task duration to the time required for each step of the online solver which is conservatively approximated by the offline solve time. For example, in the 30 task case, the offline solver takes roughly 40 seconds---it would be applicable in missions where task durations are greater than a minute. Otherwise, we could modify the online solver to run every two steps, for example.
\end{remark}

\subsection{Stochastic Task Allocation Experiments}
\label{sec:stochastic-mission-models}
We evaluate our online solution framework on a testbed that generates random task graphs and simulates uncertainty in task rewards. We compare the offline flow-based (Offline) and online flow-based (Online) task allocation algorithms in experiments that use two distinct methods of representing model error. The first,  \textit{stochastic task failure}, represents when a task fails entirely, e.g., agents are blocked from moving down a path necessary for a transportation task. The second, \textit{mission model uncertainties}, represents broad inaccuracies in the reward model, as might occur with changing conditions or under-calibrated task reward models. 

\textbf{Stochastic Task Failure}%

The stochastic task failure model produces task failures randomly with a specified likelihood $p_f \in [0,1]$, where failure means that the task returns zero reward. This results in a reward model that is perfectly accurate when tasks succeed, but which fails to predict occasional task failures.


We implement a class of baseline solvers termed \textit{Clairvoyant Solvers}: these solvers experience task failures, but \textit{have advance knowledge of which tasks will fail} via a ground truth reward model that returns zero rewards for these tasks. The tasks that will fail are selected randomly at trial initialization and are held constant for all solvers. The \textit{Offline Clairvoyant Solver (Clair-Off)} applies the offline flow-based task allocation method in an open-loop form. The \textit{Online Clairvoyant Solver (Clair-On)} applies the online flow-based task allocation method, re-planning after each task is completed. Although clairvoyant solvers experience the impacts of task failure, they are able to plan around these failures due to their advance knowledge, whereas the Online solver must react to task failures as they happen. The clairvoyant solvers thus represent a better-than-possible ideal of how solvers could react to task failures.

\textbf{Mission Modeling Uncertainties:}
To represent modeling uncertainties, we perturb each parameter in the ground truth reward model to produce an approximate perturbed reward model. The model parameters that are perturbed are coalition function parameters and influence function parameters, and define the specific instance of the function given its general form. For example, we use the sigmoid function for representing both coalition and influence functions:
\[r = \frac{c_0}{1 + e^{-c_1(x- c_2)}}, \]
where, $x$ is the function input (the coalition fraction when used as a coalition function, or the prior task reward when used as an influence function), and $\{c_0, c_1, c_2\}$ is the set of function parameters, which control the maximum value, slope, and position of the step, respectively. 

To perturb each model parameter, we choose a new value from a random distribution centered at the ground truth value: $\tilde{c} \sim \mc{N}(\mu = c, \sigma = pc)$, where $c$ is the true parameter value, and $p_m \in [0,1]$ is an error factor that determines the standard deviation as a scalar multiple of the mean. This operation is performed with all coalition function and influence function parameters for each task, using the same fixed error scale value $p_m$, and results in model inaccuracy.  

We use the Clairvoyant class of solvers (Clair-Off and Clair-On) as baselines for the model uncertainty experiments. These clairvoyant solvers have access to the ground truth reward model---instead of knowing in advance of task failures, they know the true reward model parameters, and thus can predict the reward of a given solution perfectly.

\subsubsection{Experiment 7: Stochastic Task Failure}
In this experiment, we use the stochastic task failure error model to evaluate the performance of the Offline and Online solvers when tasks fail. We compare to two baselines: the clairvoyant offline and online solvers that have advance knowledge of which tasks will fail (Clair-Off, Clair-On). 

\begin{figure}[t]
	\centering 
	\includegraphics[trim={0.0cm 1.0cm 0.0cm 0.0cm},clip,width=\linewidth]{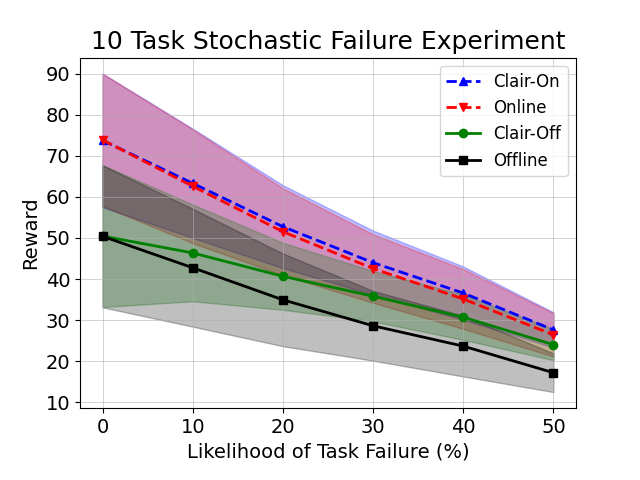}
 	\includegraphics[trim={0.0cm 1.0cm 0.0cm 0.0cm},clip,width=\linewidth]{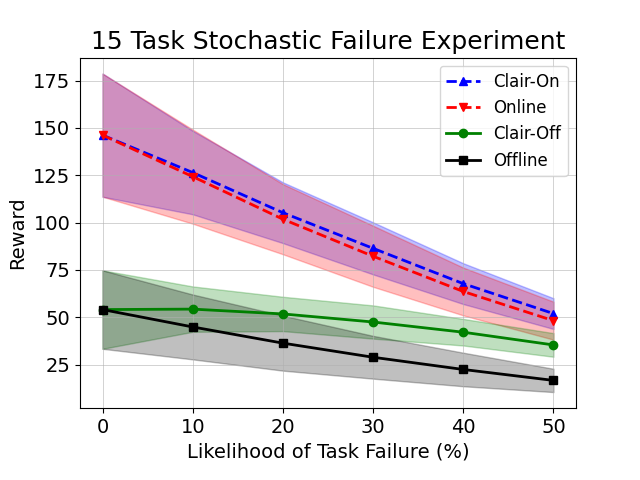}
	\caption{\label{fig:catastrophic-10} This experiment compares the online and offline solvers against the clairvoyant baselines. We vary the likelihood that any given task will fail (return zero reward) from 0 to 50\%. (top) shows missions with 10 tasks and (bottom) shows missions with 15 tasks.}
\end{figure}

We evaluate the relative performance of our four solvers as we vary the likelihood of task failure from $p_f = 0\%$ to $50\%$. We show results for trials with 10 tasks and 15 tasks in \fgref{fig:catastrophic-10}. For each of these task quantities, we generate 100 random missions for each value of $p_f$ in $[0\%, 10\%, 20\%, 30\%, 40\%, 50\% ] $. For each distinct mission, we draw task failures randomly 50 times, run all solvers, and average the results. The figures show the mean reward values for each solver averaged over the 100 trials, and the shaded areas represent the standard deviation of the reward values over the 100 trials. In both the 10-task and 15-task experiments, the online solvers perform significantly better than the offline solvers. For the online solvers, the gap between the actual and clairvoyant performance stays constant as failure likelihood increases, and the gap is relatively small---on the order of $10-20\%$ of the absolute reward values. This indicates that the online solver adapts well even as failures increase to $50\%$. This is due to the ability of the online solvers to incorporate the feedback from failed tasks -- at each solver iteration, the observed rewards are integrated into the updated reward model, allowing the online solvers to re-allocate agents away from tasks whose performance would be compromised due to influence from the failed task. For the offline solvers, we see lower overall performance, and the gap between the clairvoyant and the actual solver performance grows to nearly $50\%$, indicating an inability to recover from task failures.

\subsubsection{Experiment 8: Mission Modeling Uncertainty}
\begin{figure*}
	\centering 
	\includegraphics[trim={0.2cm 0.0cm 1.5cm 0.0cm},clip,width=0.32\linewidth]{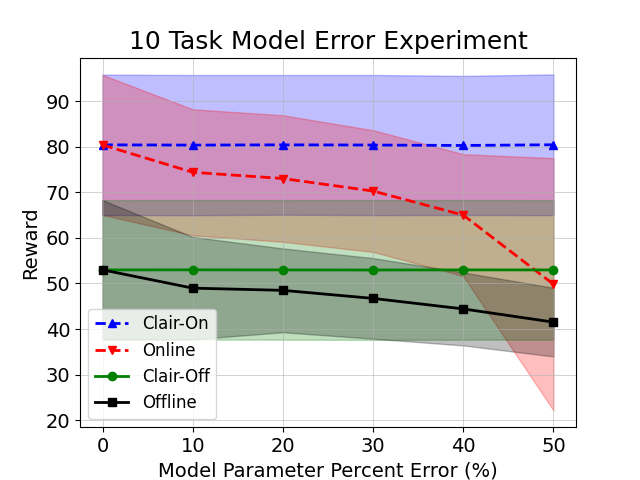}
 	\includegraphics[trim={0.2cm 0.0cm 1.5cm 0.0cm},clip,width=0.32\linewidth]{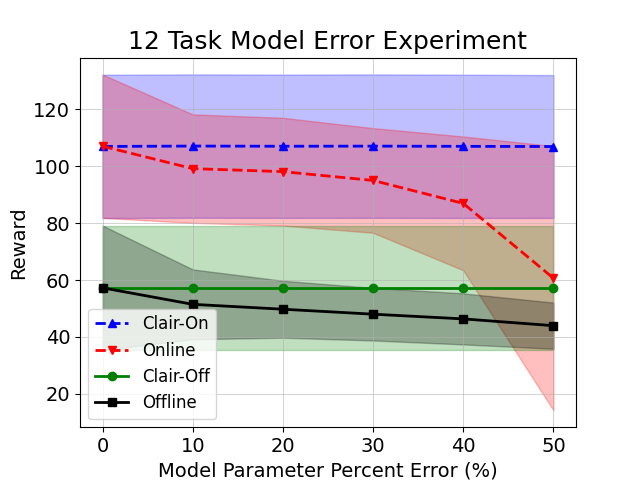}
  	\includegraphics[trim={0.2cm 0.0cm 1.5cm 0.0cm},clip,width=0.32\linewidth]{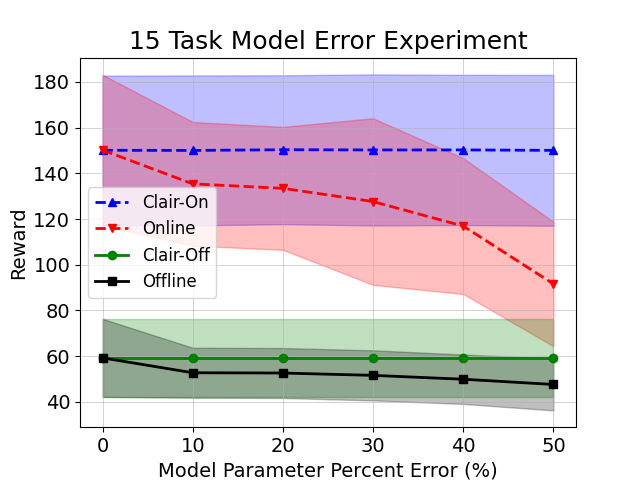}
	\caption{\label{fig:model-10} This experiment compares the average performance of the online and offline solvers against the clairvoyant baseline solvers as the percent error of reward model parameters is varied from 0\% to 50\%. The plots show the average reward for each of the solvers on randomly generated with 10, 12, and 15 tasks, from left to right. The shading shows the standard deviation of each solver's performance.}
\end{figure*}

In the model uncertainty experiments, we compare the performance of plans generated by the offline and online flow-based solvers (Offline, Online) given an incorrect system model with the performance of clairvoyant solvers given access to the ground truth model (Clair-Off, Clair-On). We test model error parameters of $p_m = [0\%, 10\%, 20\%, 30\%, 40\%, 50\% ] $. At each parameter value, we generate 100 distinct ground truth mission models. For each of these ground truth models, we draw 50 distinct perturbed models. We independently evaluate both the offline flow-based solver and the online solver on experimental runs planned using each of the 50 models -- the model creates task allocation plans using the perturbed model, and observes rewards calculated using the ground truth model. We average these results together to yield average performance for each trial. The figures show the average reward over all 100 trials for each level of model perturbation for each of the four solvers. The error shading on the figures indicates the standard deviation of the rewards over the 100 trials. \fgref{fig:model-10} shows these results for missions with 10, 12, and 15 tasks. 

The results show that the online flow-based solver is effective at adapting the task plan, even when given a significantly perturbed mission model.
As the perturbation level increases, performance degrades in both the online and offline solvers compared to the clairvoyant versions.
The relative degree of degradation is similar between the online and offline solvers.
The online solver with the perturbed model (Online) outperforms the clairvoyant offline solver (Clair-Off) across the domain of perturbation, only drawing even when $p = 50\%$ and perturbed models are highly inaccurate.
This is true despite the models with large perturbations becoming numerically unstable in some instances, a phenomenon partly responsible for the large variance in some  trials.

\section{Advanced Task Allocation Experiments}\label{sec:advanced-experiments}
\begin{figure}[htbp]
	\centering 
	\includegraphics[trim={0.0cm 4.0cm 18.8cm 1.0cm},clip,width=0.9\columnwidth]{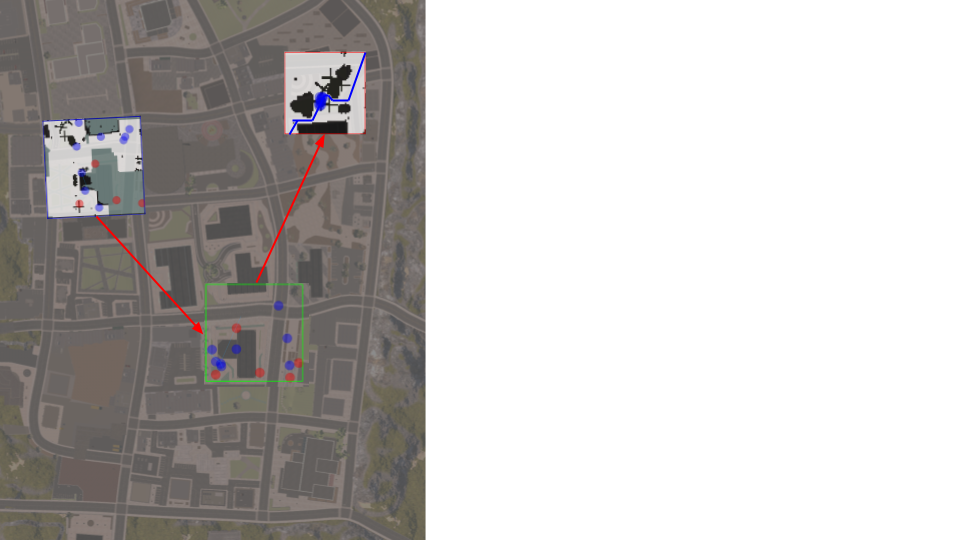}
	\caption{\label{fig:dcist-testbed} A planview of the advanced testbed shows buildings and street features. The agents (blue dots) in the top left work on an exploration task; the agents in the center work on coverage control of features of interest (red dots); the agents in the top right work on a transport task, with their selected path shown in blue. The grids shown in the exploration and transport task are the occupancy grid at the ground level.}
\end{figure}

In real-world tasks, the reward is determined by a measurable objective function of the environment state, e.g., the size of a region explored or the quantity of a material transported.
However, our previous experiments evaluated a task allocation plan by applying our \textit{model} of task reward described in~\scref{sec:mission-model}; this model is only an estimate.
The true reward depends upon many more factors than the number of robots completing the task and the prior task rewards---it may be impacted by wind speed, obstacles, and unexpected robot interactions.
To evaluate the impact of such environmental factors on task planning accuracy and performance, we implement a high-fidelity testbed capable of simulating the kinematics of multi-robot teams completing many tasks over a large (3200mx3200m) urban environment with obstacles (see~\fgref{fig:dcist-testbed}).
The simulator has a hierarchical architecture: an occupancy grid and robot states and actions are processed by detailed models for sensing, communication, and kinematics, and rendered with Unity.
It is capable of simulating large (100+) robot teams operating on parallel tasks in real time by leveraging communication in ROS.
Docker-Compose isolates the individual processes of each agent from one another.  
The testbed has been used to simulate inter-robot communication problems~\cite{MOX,cladera2023enabling}, exploration and navigation problems with specific attention to minimizing the sim-to-real gap~\cite{cai2022risk, cladera2024challenges}, and multi-agent task allocation approaches~\cite{messing2023sampling, banfi2022hierarchical}.
In our implementation, we create controllers for four example tasks and design a graph manager that manages planning, agent assignment, and inter-task travel.
These tasks each have reward outputs grounded in physical phenomena in the simulator, enhancing the ability to represent real-world uncertainty in task modeling and execution.
The simulator runs on a System76 desktop equipped with an AMD Ryzen Threadripper, 64 GB of RAM, and an NVIDIA RTX 2070 Super GPU.

\subsubsection{Task Types, Controllers, and Reward Modeling}

To represent a wide variety of missions that might occur in agriculture, construction, or emergency management scenarios, we implemented four representative classes of tasks: coverage control, exploration, multi-agent transportation, and multi-agent collaborative carry. These four tasks represent a broader variety of real robot tasks: for example, coverage control can be a stand-in for monitoring pests in a field, following hot spots of a forest fire, or monitoring areas where an injured person in need of rescue may be located. These tasks also express the different types of reward functions that are present in real-world tasks:
for example, an exploration task has a coalition function that is sublinear in the number of agents, whereas a collaborative carry task has a step-like coalition function.

Each task has a reward value that is measurable based on physical phenomena in the simulator: for coverage control---the coverage cost calculated from agent positions and feature of interest positions; for the exploration task---the region that has been explored at the end of the task; and for the transportation and collaborative carry tasks---the quantity of resources transported.
Because we ground the reward observations in physical metrics, we can compare our reward model estimates to a ground truth that accounts for many more of the complexities present in executing multi-robot tasks. 
According to our mission assumptions, each task has a pre-specified fixed duration. We estimate inter-task travel time with the Euclidean distance between two tasks and agent maximum speed. 
We developed a reward model for each task through a combination of expert design (choosing the form of functions, e.g., step function or sublinear) and experimentation to fine-tune function coefficients.

\textbf{Exploration: } The objective is to explore a specified region of interest, represented by an occupancy grid. We apply a centralized, frontier-based approach~\cite{Yamauchi1997}---agents explore unoccupied cells, avoiding collisions, and report their discoveries to a central broker.
We use the flood fill algorithm~\cite{burtsev1993efficient} for frontier detection, and A* ~\cite{hart1968} for frontier selection. 
We model agent sensing with a square sensor that perfectly observes occupied vs.\ free cells.
Additionally, ray-tracing is performed to ensure agents cannot see behind obstacles.

The exploration reward is defined as the percentage of the unoccupied environment cells discovered during the fixed task duration: $\frac{\text{\# unoccupied cells explored}}{\text{\# total unoccupied cells}}$.

\textbf{Coverage Control: }
Given a team of robots, a set of features of interest, and a target region, the coverage control task generates agent actions that result in even coverage of the features of interest, minimizing the \textit{coverage cost} metric~\cite{cortes2004coverage}, which serves as the observable reward of the task. The coverage cost measures the importance-weighted distance of each robot to its nearest regions of interest to which it is nearest
In our implementation, we used a linear function to translate this cost into a reward. 

Static ``feature'' robots serve as landmarks that the task robots must monitor.
We use a centralized implementation of Lloyd's algorithm \cite{lloyd1982least}.
The task ends after its fixed duration, and the final coverage reward is reported.

\textbf{Multi-Agent Transportation: }
The objective of the multi-agent transportation task is to move objects from a designated depot to a specified goal location. Each robot can carry one unit at a time, and must avoid obstacles and other robots. 
The reward is the sum of units delivered to the goal location. The total units available at the start depot (i.e., the maximum possible task reward) depends on the performance of preceding tasks.

The transport controller moves agents along paths planned on an occupancy grid representation of the task region. The controller plans paths independently for each agent using the Local Repair A* (LRA*) algorithm~\cite{silver2005cooperative}.
Conflicts are resolved heuristically. If the task duration is sufficiently long, the controller will plan multiple trips.

After a robot completes all planned trips, it redirects to a random location to avoid blocking the goal location. The task is terminated after its fixed duration and the number of resources at the goal is reported as the reward.

\textbf{Collaborative Carry Transportation: }
In this task, agents must cooperate to carry larger objects with greater value. Objects have ``weights'' corresponding to the number of robots required to carry them.
The reward earned by the collaborative carry task is a sum of values of objects delivered. The values of objects are a function of object weight and other factors determined by performance of prior tasks. 

The collaborative controller functions largely the same as the multi-agent transport controller, but instead of planning paths for individual agents, the controller plans the paths and commands the actions of multi-agent collectives of robots.
These collectives are established with the fixed relative positions of a group of robots as they pick up an object, constraining their relative positions until the object is delivered to the goal depot.
If enough objects exist at the start depot, multiple teams can transport objects simultaneously while avoiding collisions, and robots may make multiple trips. 

\subsubsection{Graph Manager}
The graph manager module is a meta-controller that manages robot deployment, inter-task travel, task feedback, and task planners. Initialized with a mission specification and task graph model, it handles task types, locations, durations, and task-specific parameters like feature count and object weights.

Using ROS, the graph manager oversees task execution and feedback. Upon mission start, it launches ROS nodes for each task, positions robots randomly in the simulator, and initializes the specified solver (offline or online). When agents are assigned, the controller directs them to the task regions and switches control to a task-specific controller upon arrival.

After task completion, the ROS action server reports task rewards to the graph manager, which assigns the next task once the agents are available. If using the online solver, the manager updates the task plan after each task based on progress, rewards, free agents, and current time. 

The graph manager ends the mission when no tasks remain or the makespan constraint is reached, recording observed task rewards.

\subsubsection{Reward Modeling of the Implemented Tasks}
\label{sec:advanced-reward-modeling}
To apply any of the solvers developed in this work, we require a precise specification of the mission: the graph topology and the reward model. We randomly generate graph topology
and leave specific graph topologies to case studies (see~\scref{sec:case-study}).

In order to develop accurate reward models for all tasks and inter-task relationships, we use a mixture of expert design and empirical evidence from task experiments. For an illustrative example, we examine the development of the reward model for the multi-agent transport task. First, we conducted 500 trials of the multi-agent transport task, initialized with random robot locations and a unique task duration $d_t$, region size, occupancy grid, and number of robots. 
Using these results, we identify an overall form for the coalition function 
$\rho(C_T) = a_{0}+a_{1}\frac{C_T}{N},$
where $C_T$ is the number of agents assigned to the transport task and $\{a_1, a_2\}$ are non-negative scalars that represent the set of function parameters. The function parameters are functions of task properties such as duration, region size, and occupancy grid density. A linear function was chosen as it best represents the relationship between the number of agents and the amount of resources transported---each robot carries one unit, and we found that inter-robot interference effects were marginal. The parameter functions $\{a_0, a_1,\}$ were then fine-tuned to minimize the prediction error of the reward model over the experimental dataset.

The performance on tasks preceding the transport task determines the total number of resources available at the start depot. We approximate this relationship with another linear function:
$\delta_{iT}(r_i) = a_{2,i} r_i + a_{3,i},$
where $\{a_{2,i}, a_{3,i}\}$ are scalar functions of properties of the preceding task $i$. We select $\mc{A}$, the influence aggregation function, to be the ``sum'' operation, and select $\ddagger$, the coalition-influence aggregation function, as the ``minimum'' operation. The minimum function represents the relationship between the influence (number of resources available at the start) and coalition (number of resources the robots can transport in the task duration). 
A full set of reward model specifications for all tasks is in Table~\ref{tab:reward-model}.

\begin{table*}[t]
  \renewcommand{\arraystretch}{1.3}
    \centering
    \caption{Reward models for four tasks in the advanced testbed    \label{tab:reward-model}}
\begin{tabular}{ccccc}
         & Coverage & Exploration & MA Transport & Collab. Carry \\
         \midrule
        Coalition Function ($\rho_j$) & $a_{cov,0} - a_{cov,1}\frac{C_T}{N}$ & $a_{e,0}(1 - e^{-a_{e,1} \left(\frac{C_T}{N}\right)})$ & $a_{t,0}+a_{t,1}\frac{C_T}{N}$ & $\frac{a_{coll,0}}{1+e^{-a_{coll,1}(\frac{C_T}{N} - a_{coll,2}})}$\\
        Influence Function($\delta_{ij}$) & $\frac{1}{|\Nin[j]|}r_i$& $\frac{1}{|\Nin[j]|}r_i$ & $\frac{1}{|\Nin[j]|}r_i$ & $\frac{1}{|\Nin[j]|}r_i$\\
        Influence Agg. Function($\mc{A}_j$) & Sum & Sum & Sum & Sum \\
        Coalition-Influence Agg. Function($\ddagger_j$) & Product & Product & Minimum & Product \\
        \bottomrule
    \end{tabular}
\vspace{-0.5cm}
\end{table*}

\begin{table}[htbp]
\renewcommand{\arraystretch}{1.3}
    \centering
    \caption{Symbols and definitions for advanced testbed reward model}
    \begin{tabular}{p{0.2\linewidth} | p{0.6\linewidth}}
        Symbol &  Meaning\\
        \hline
         $a_{cov,0}$,$a_{cov,1}$ & Coverage parameters, dependent upon region size and number of features  \\
         $a_{e,0}$ & Max. possible exploration reward \\
         $a_{e,1}$ & Exploration diminishing returns slope \\
         $a_{t,0}$ & Max. possible multi-agent transport reward \\
         $a_{t,1}$ & Multi-agent transport diminishing returns slope \\
         $a_{coll,0}$ & Collaborative carry max. reward \\
         $a_{coll,1}$ & Sigmoid slope \\
         $a_{coll,2}$ & Sigmoid step placement along x-axis  
         
    \end{tabular}
    \label{tab:my_label}
\end{table}

\subsection{Advanced Testbed Case Study}\label{sec:case-study}
This section presents a case study to \textit{(i)}~illustrate the modeling of tasks and their rewards, \textit{(ii)}~exemplify differences in the online and offline solvers, and \textit{(iii)}~demonstrate the execution of the overall mission.

\begin{figure}[ht]
	\centering 
	\includegraphics[width=0.9\columnwidth]{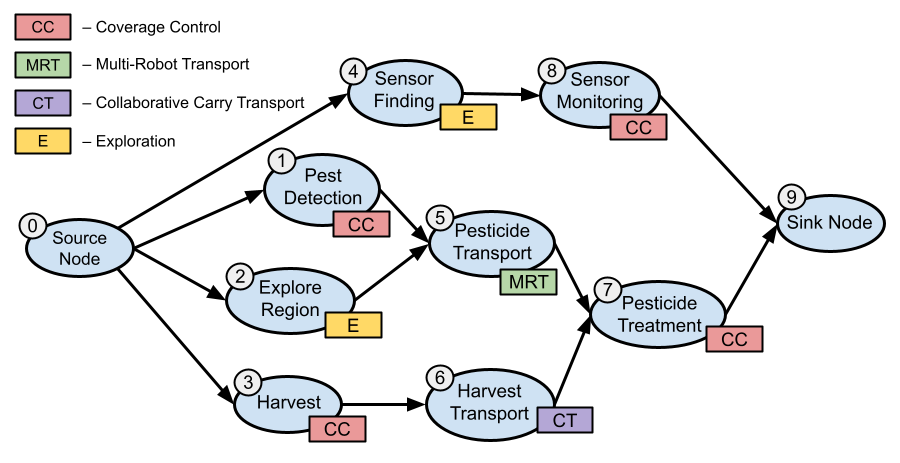}
	\caption{\label{fig:case-study-graph} A task graph representation of the agriculture case study.}
\end{figure}

A team of ten agents is deployed to execute an autonomous agriculture mission, modeled by the graph shown in~\fgref{fig:case-study-graph}. In the mission, crops must be harvested to make way for pesticide treatment in a pest-infested area. In order to accomplish this, the pests must be characterized, pesticide fluid transported through an unmapped area to the infested site, and crops harvested and carried away to avoid getting sprayed. Additionally in a separate field, distributed soil sensors must be located and monitored. We note in~\fgref{fig:case-study-graph} the task type used to represent each task in the case study. 
\Cref{tab:reward-model} gives the reward functions for all task types in the case study. 

\subsubsection{Offline Solution}
The offline solution generates a straightforward task plan: allocate all 10 agents to tasks 4, 8, and 9, as shown in pane 0 within~\fgref{fig:case-study-solns}. This solution is locally optimal for the offline algorithm because assigning agents to tasks on the lower branches requires dividing agents among several branches. Due to the structure of the mission, small coalitions are expected to perform poorly on the tasks in the lower branches. For example in task 6, harvest transport, 7 agents are required to successfully carry the resources. The offline solver's solution rewards for each task are listed in \Cref{tab:offline-case-study}---the offline solver's solution earns a reward of 77.9 after execution. 

\subsubsection{Online Solution}
\begin{figure}[t]
	\centering 
	\includegraphics[trim={1.7cm 0.0cm 2.4cm 0.0cm},clip,width=0.47\linewidth]{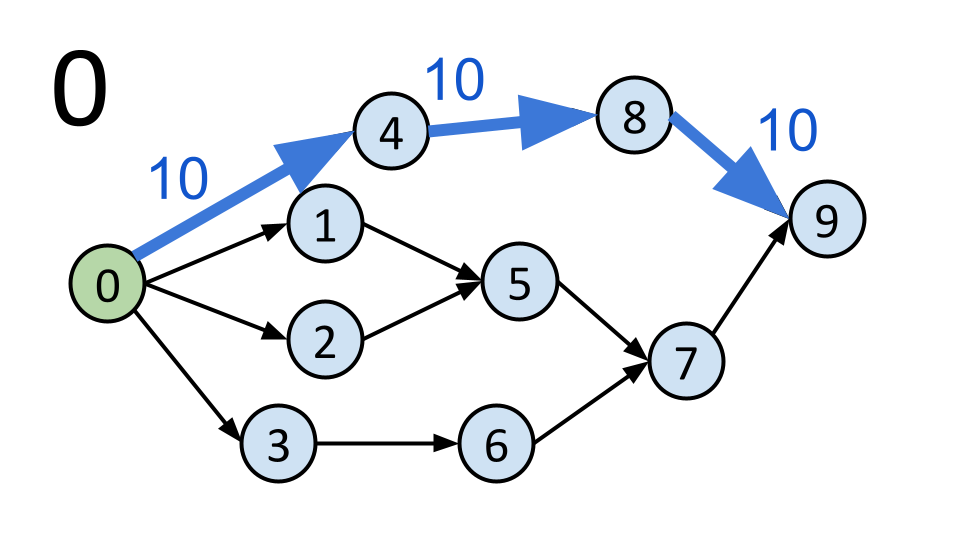}
 	\includegraphics[trim={1.7cm 0.0cm 2.4cm 0.0cm},clip,width=0.47\linewidth]{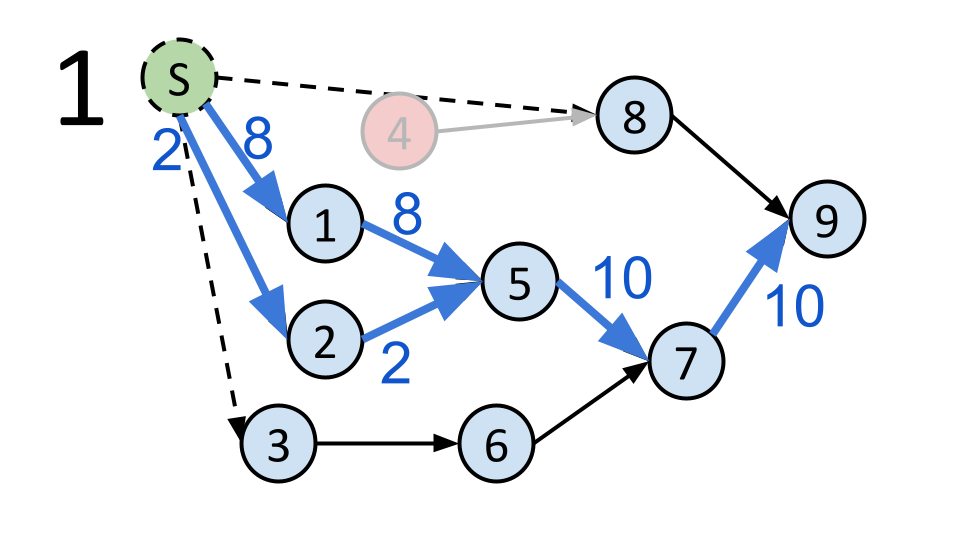}
  	\includegraphics[trim={1.7cm 0.0cm 2.4cm 0.0cm},clip,width=0.47\linewidth]{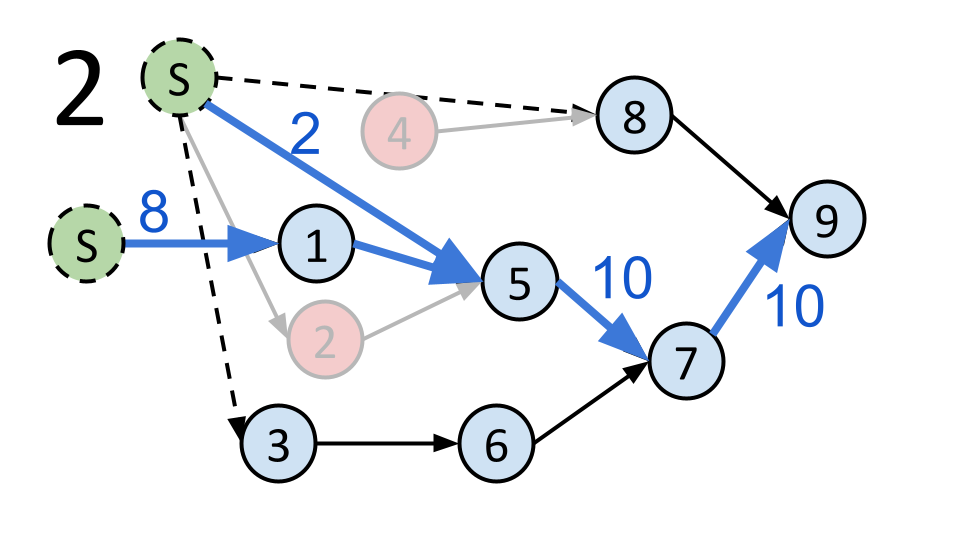}
 	\includegraphics[trim={1.7cm 0.0cm 2.4cm 0.0cm},clip,width=0.47\linewidth]{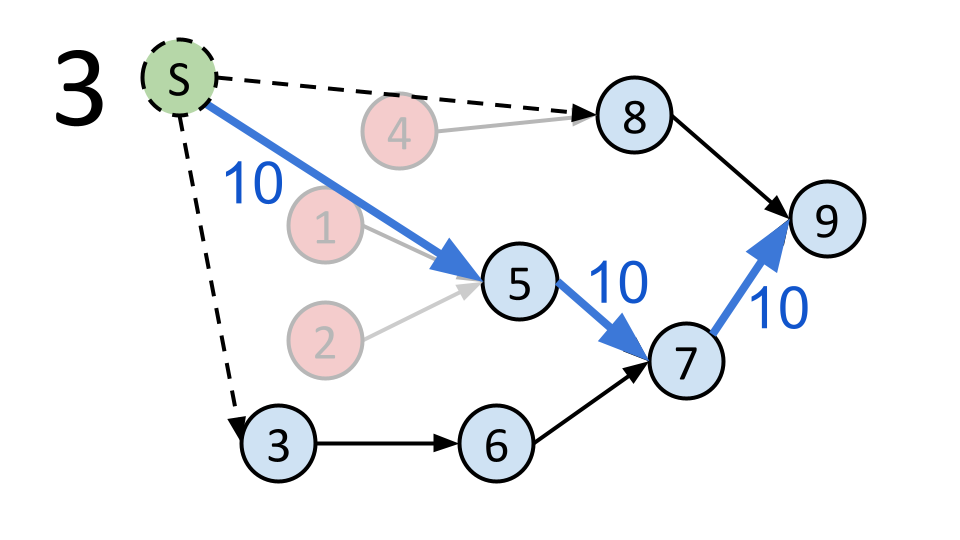}
   	\includegraphics[trim={1.7cm 0.0cm 2.4cm 0.0cm},clip,width=0.47\linewidth]{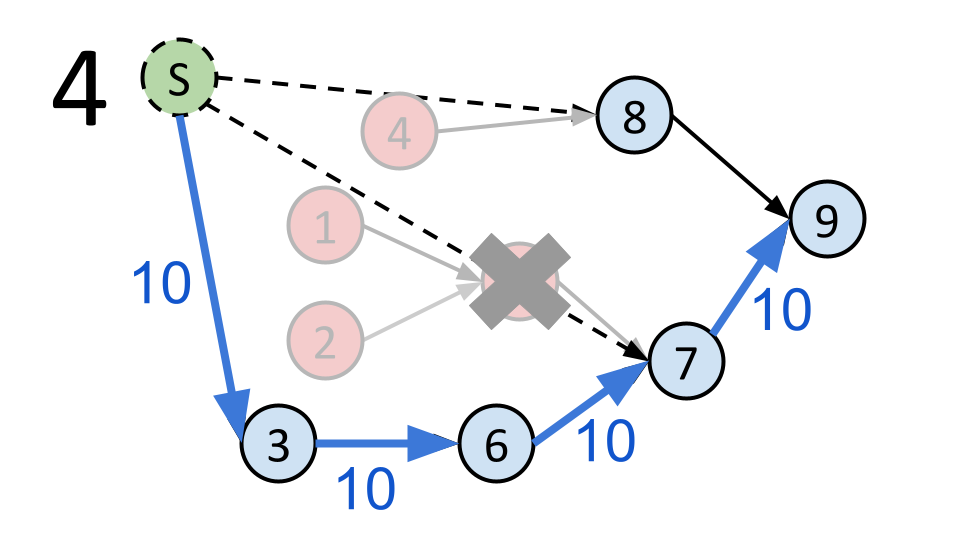}
   	\includegraphics[trim={1.7cm 0.0cm 2.4cm 0.0cm},clip,width=0.47\linewidth]{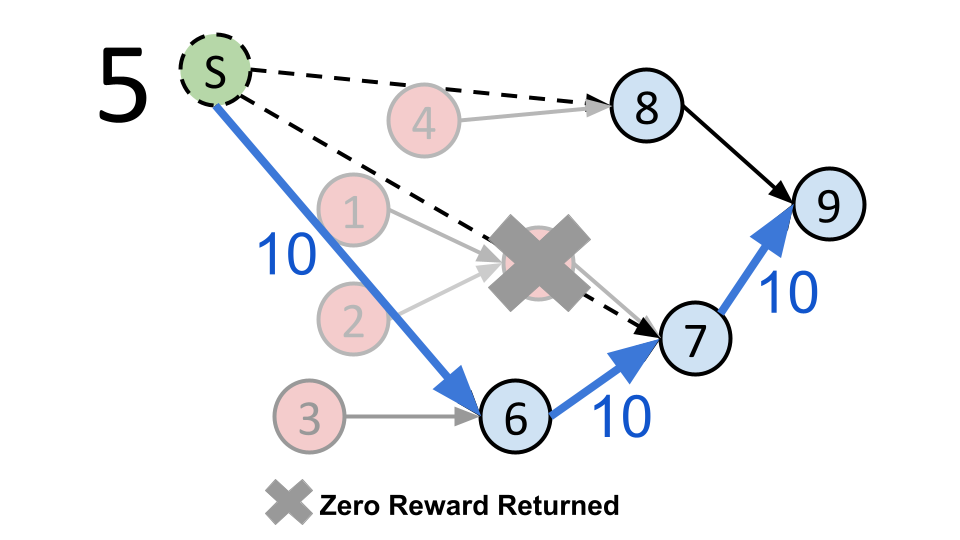}
	\caption{\label{fig:case-study-solns} Flow-based solutions of task graph in \fgref{fig:case-study-graph}. (Step 0) The offline solution allocates all 10 agents along the top branch. (1) After completing task 4, the online solution adds the artificial source node marked ``S'' and the new dashed edges. The new solution is shown, dividing agents among three branches. Modified task graphs and corresponding flow solutions are shown for the first 5 steps. Completed nodes are marked in red, and source nodes are marked in green and labeled with an ``S''---an additional artificial source node is created in step 2 to feed into the in-progress task 1. The number of agents allocated to flow along each edge is indicated in blue.}
    \vspace{-0.5cm}
\end{figure}

The online solver begins with the same solution as the offline solver, but iteratively re-computes the flow solution as tasks are completed. In step 1 of~\fgref{fig:case-study-solns}, agents first complete task 4, and earn a reward 14\% lower than expected. After completing task 4, the online solver's step function is called: the graph is restructured to remove task 4, create an artificial source node with all 10 agents, and connect this artificial source node to tasks 1, 2, 3, and 8, thereby enabling the agents to flow across branches of the task graph. The reward model is updated with task 4's observed reward. On the updated task graph, the solver is run again, and a new plan selected: now, agents are divided between tasks 1 and 2, reconvene at task 5, and then finish the mission by completing task 7. In pane 4 of~\fgref{fig:case-study-graph}, we see that task 5 fails. 
The online solver is run again with this observation (zero task reward) reflected in its reward model. The revised solution redirects flow to the lower branch of the graph in response, completing tasks 3 and 6 in order to make completing task 7 worthwhile. The online solution described here resulted in an increase of 29\% in reward over the offline solution. The rewards for each task in completion order are listed in \Cref{tab:offline-case-study}. 

\begin{table}[ht]
\renewcommand{\arraystretch}{1.2}
    \centering
    \caption{Case Study Task Rewards\vspace{-0.5cm}}
    \subfloat[Offline Solver]{
    \begin{tabular}[b]{crr}
        Task & Expected & Actual \\
        \toprule
        4 & 49.7 & 42.9 \\
        8 & 28.2 & 32.3 \\
        \midrule
        TOTAL & 77.9 & 75.2
    \end{tabular}
    }
    \subfloat[Online Solver]{
    \begin{tabular}[b]{crr}
        Task & Expected & Actual \\
        \toprule
        4 & 49.7 & 42.7 \\
        1 & 6.3 & 7.6 \\
        2 & 3.2 & 4.3 \\
        5 & 30.0 & \textcolor{red}{0.0*} \\
        3 & 7.8 & 8.4 \\
        6 & 24.0 & 25.2  \\
        7 & 9.8 & 8.6  \\  
        \midrule
        TOTAL & 130.8 & 96.8
    \end{tabular}   
    }    
    \label{tab:offline-case-study}
\end{table}

\subsection{Experiments in Advanced Testbed}
We test the online and offline solvers in the advanced testbed through a series of randomly generated missions.
Given a number of tasks, number of agents, graph topology parameters, and a makespan constraint, we randomly generate mission specifications comprising the following: graph topology, task types, task locations, task region sizes, task durations, and individual task parameters such as quantity and location of features of interest in the coverage control task, or the weight of items transported in the collaborative carry task.
For each  mission, we run independent evaluations for both offline and online solver plans.
We record expected and actual reward values for each of the tasks and for the entire mission, and use the overall reward metrics to compare the two solvers.

\subsubsection{Reward Model Accuracy}
In \Cref{table:dcist-reward-model-accuracy}, we show the average actual reward, average expected reward, and average percent error, calculated over all task instances of the 92 trials.
The actual reward values are from task outputs. The expected reward values are the reward model outputs given accurate coalition and prior task reward data.
The average percent error is the absolute value of the percent error for each task execution and averaging over all task instances. 

The coverage control, exploration, and multi-agent transport tasks each have average errors between $20\%$ and $25\%$, and the collaborative carry task has an average error of just $7.7\%$.
These error values are well within the error margins tested in Experiment 7, wherein the online solver performance degraded minimally ($10\%$) compared to the clairvoyant task plans.
This demonstrates that, when feedback is incorporated via the online solver, even relatively simple error models are able to model tasks sufficiently for high-quality task plans.

\begin{table}
\renewcommand{\arraystretch}{1.2}
  \centering
  \caption{Reward Model Accuracy Data}%
  \label{table:dcist-reward-model-accuracy}
  \begin{tabular}{r|r|r|r}
        Task Type & Actual $r$ & Expected $r$ & Avg. Abs. Error \\
        \hline
        Coverage Control & 4.60 & 4.36 & 24.3\% \\
        Exploration & 5.43 & 5.93 & 21.9\% \\
        Multi-Agent Transport & 2.22 & 2.43 & 22.8\% \\
        Collaborative Carry & 5.52 & 5.20 & 7.7\% \\
        
  \end{tabular}
\end{table}

\subsubsection{Experiment 9: Offline vs. Online Solvers in Advanced Testbed}
\begin{figure}[tbp]
	\centering 
	\includegraphics[trim={0.0cm 0.0cm 0.0cm 0.0cm},clip,width=\linewidth]{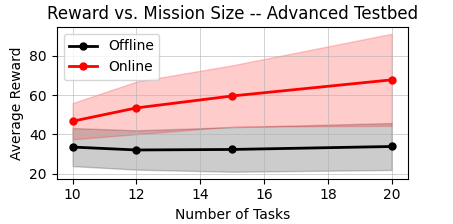}
	\caption{\label{fig:n-tasks-advanced} We show average reward vs number of tasks in the mission for experiments consisting of 25 trials in the advanced testbed. The shading shows the standard deviation for each solver.}
\end{figure}

We run multiple random trials for mission sizes of 10, 12, 15, and 20 tasks and report average reward values over the entire mission for both the online and offline solvers (\fgref{fig:n-tasks-advanced}).
The online solver outperforms the offline solver by 30\% in the 10 task missions, growing to 100\% for 15 task missions.
The online solver's performance improves relative to the offline solver as mission size increases.
This mirrors the trend seen in the simplified testbed experiments (\scref{sec:online-experiments-simplified}). 

\subsubsection{Experiment 10: Average Reward vs Team Size in Advanced Testbed}
\begin{figure}[tbp]
	\centering 
	\includegraphics[trim={0.0cm 0.3cm 0.0cm 0.0cm},clip,width=\linewidth]{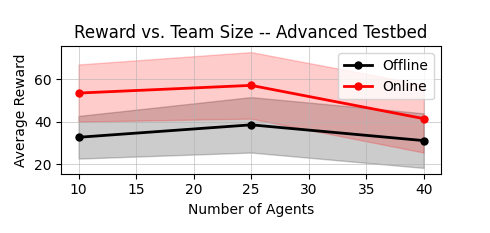}
	\caption{\label{fig:n-team-advanced} We show average reward vs number of agents within the team accomplishing a mission for experiments consisting of 25 trials in the advanced testbed. The shading shows the standard deviation for each solver.}
\end{figure}

We run multiple random trials for team sizes of 10, 25, and 40 agents and report average reward values over the entire mission for both the online and offline solvers (\fgref{fig:n-team-advanced}). The online solver outperforms the offline solver by 63\% in the 10 agent missions, by 48\% in the 25 agent missions, and by 33\% in the 40 agent missions. The performance of both solvers decreases with larger team sizes because the reward model predictions are poorer. The reward models for the advanced testbed were calibrated with 10 agents. Average error increases to greater than 50\% for all tasks on trials with 40 agents.

\subsubsection{Experiment 11: Average Reward vs Catastrophic Error in Advanced Testbed}
\begin{figure}[tbp]
	\centering 
	\includegraphics[trim={0.0cm 0.3cm 0.0cm 0.0cm},clip,width=0.9\linewidth]{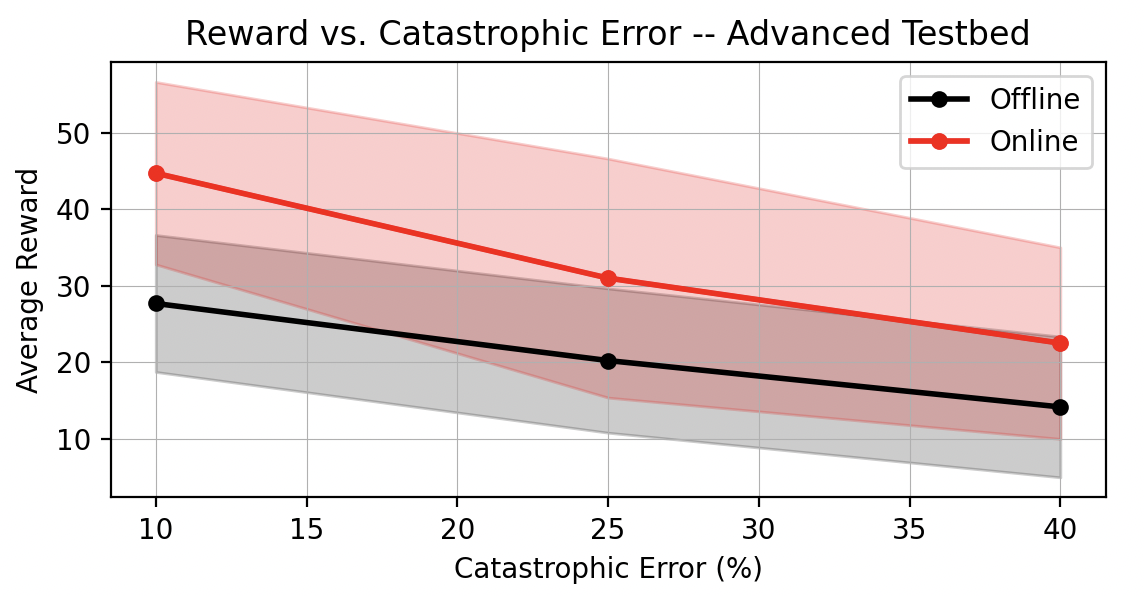}
	\caption{\label{fig:cat-error-advanced} We show average reward vs probability of catastrophic task failure rate within the team accomplishing a mission for experiments consisting of 25 trials in the advanced testbed. The shading shows the standard deviation for each solver.}
\end{figure}

We conduct multiple random trials for catastrophic task failures with probabilities of 10\%, 25\%, and 40\%, reporting the average reward values over the entire mission for both online and offline solvers (\fgref{fig:cat-error-advanced}).The online solver outperforms the offline solver by between 55\% and 63\%. The reward models for the advanced testbed were calibrated with 10 agents and 12 task missions. 

\section{Conclusion and Future Work}
This article introduced the task graph mission model, which enables the representation of inter-task relationships with a rich function space and the modeling of the effects of multi-robot coordination and collaboration.
A flow-based task allocation formulation, agnostic to the number of agents, was developed. This representation was leveraged to design novel offline and online algorithms.
Extensive simulations on a simplified testbed established that (1)~the flow-based approaches are more than 1000 times faster than the optimal mixed-integer approaches, (2)~the task plans produced by the online solver were within 10\% of the optimal solution in the range tested, and (3)~the approaches are tolerant to significant error.

We also tested the performance in a high-fidelity robotics simulator, which required our approach to represent complex and interrelated multi-agent tasks with rewards grounded in physical phenomena.
These experiments demonstrated that simple reward models effective at predicting task performance, and that incorporating feedback via the online flow-based approach significantly improves performance.

This work suggests several immediate extensions. First, uncertainty could be more robustly considered with explicit modeling, e.g. with a CVar cost, while remaining compatible with our modeling approach.
Additionally, several algorithm improvements could yield higher performance: augmenting the task graph with additional edges that do not represent inter-task relationships could broaden the solution domain without violating precedence relationships.

More broadly, this work unifies multi-robot coordination and cooperation across complexly interlinked tasks, scales to large teams and mission sizes, and is robust to uncertainty.
Future work should address the heterogeneous problem with a full-stack task specification, decomposition, allocation, and execution system. Such an approach would 
further bridge the gap towards real-world applications in complex and dynamic missions.

\section*{Acknowledgment}
We thank Jeremy Wang for his assistance in implementing testbed components and running experiments, and for his insightful discussion throughout the project.

\bibliographystyle{IEEEtran}

\end{document}